\documentclass[journal,twoside]{IEEEtran}
\usepackage{amsmath,amsfonts}
\usepackage{algorithmic}
\usepackage{algorithm}
\usepackage{amssymb}
\usepackage{fancyhdr}
\usepackage{array}
\usepackage[caption=false,font=footnotesize,labelfont=sf,textfont=sf]{subfig}
\usepackage{textcomp}
\usepackage{stfloats}
\usepackage{url}
\usepackage{verbatim}
\usepackage{graphicx}
\usepackage{cite}
\usepackage{booktabs}
\usepackage[font=small]{caption}
\usepackage[dvipsnames]{xcolor}
\usepackage{multirow}

\definecolor{darkgreen}{RGB}{0,100,0}

\begin{document}

\title{AnSER: Learning Ankle and Stepping Exploitation for Robust Humanoid Control based on Reduced-Order Representations}
%
\author{Yashuai Yan*$^{,1}$, Tobias Egle*$^{,2}$, Grazia Zambella$^{2}$, Christian Ott$^{2,3}$, and Dongheui Lee$^{1,3}$
\thanks{\scriptsize $^{1}$Yashuai Yan and Dongheui Lee are with the Autonomous Systems Lab, TU Wien, 1040 Vienna, Austria
		\{yashuai.yan, dongheui.lee\}@tuwien.ac.at}%
\thanks{$^{2}$Tobias Egle, Grazia Zambella and Christian Ott are with the Automation and Control Institute, TU Wien, 1040 Vienna, Austria
		\{tobias.egle, grazia.zambella, christian.ott\}@tuwien.ac.at}%
	
  \thanks{\scriptsize $^{3}$Christian Ott and Dongheui Lee are also with the Institute of Robotics and Mechatronics (DLR), German Aerospace Center, Wessling, Germany.}%
    \thanks{\scriptsize *Contributed equally to the work.}
}
\maketitle

\begin{abstract}
This paper presents \emph{AnSER}, an end-to-end reinforcement learning (RL) framework for robust bipedal locomotion that unifies footstep planning and whole-body control within a single policy. The policy jointly predicts stepping parameters, including foot placement and timing, together with leg-joint target positions. These stepping parameters are used to generate the desired foot and Divergent Component of Motion (DCM) trajectories, while a joint-level PD controller tracks the resulting motions, yielding a unified locomotion pipeline. To guide policy learning, we use the Ankle Recovery Condition (ARC), an analytical criterion derived from DCM and Center of Pressure (CoP) dynamics that identifies disturbances recoverable through CoP modulation. ARC encourages the policy to exploit ground reaction forces (GRFs) for balancing control whenever possible and to adapt foot placement and timing only when stepping becomes necessary. In addition, \emph{AnSER} optimizes the footstep planning by incorporating DCM endpoints into the reward shaping, promoting rapid recovery and convergence to the desired motion. Furthermore, AnSER uses the reduced-order representations as the robot state, comprising the robot's legs and center of mass (CoM) quantities. By regulating the robot's CoM solely through leg joints, \emph{AnSER} naturally permits arbitrary upper-body motions without requiring task-specific training. Extensive simulation and hardware experiments demonstrate that \emph{AnSER} achieves natural and robust locomotion by exploiting GRFs to reduce unnecessary stepping, while outperforming both purely model-based and end-to-end learning baselines under challenging disturbances.

\end{abstract}

\section{Introduction}
\begin{figure}
    \centering
    \includegraphics[width=0.9\linewidth]{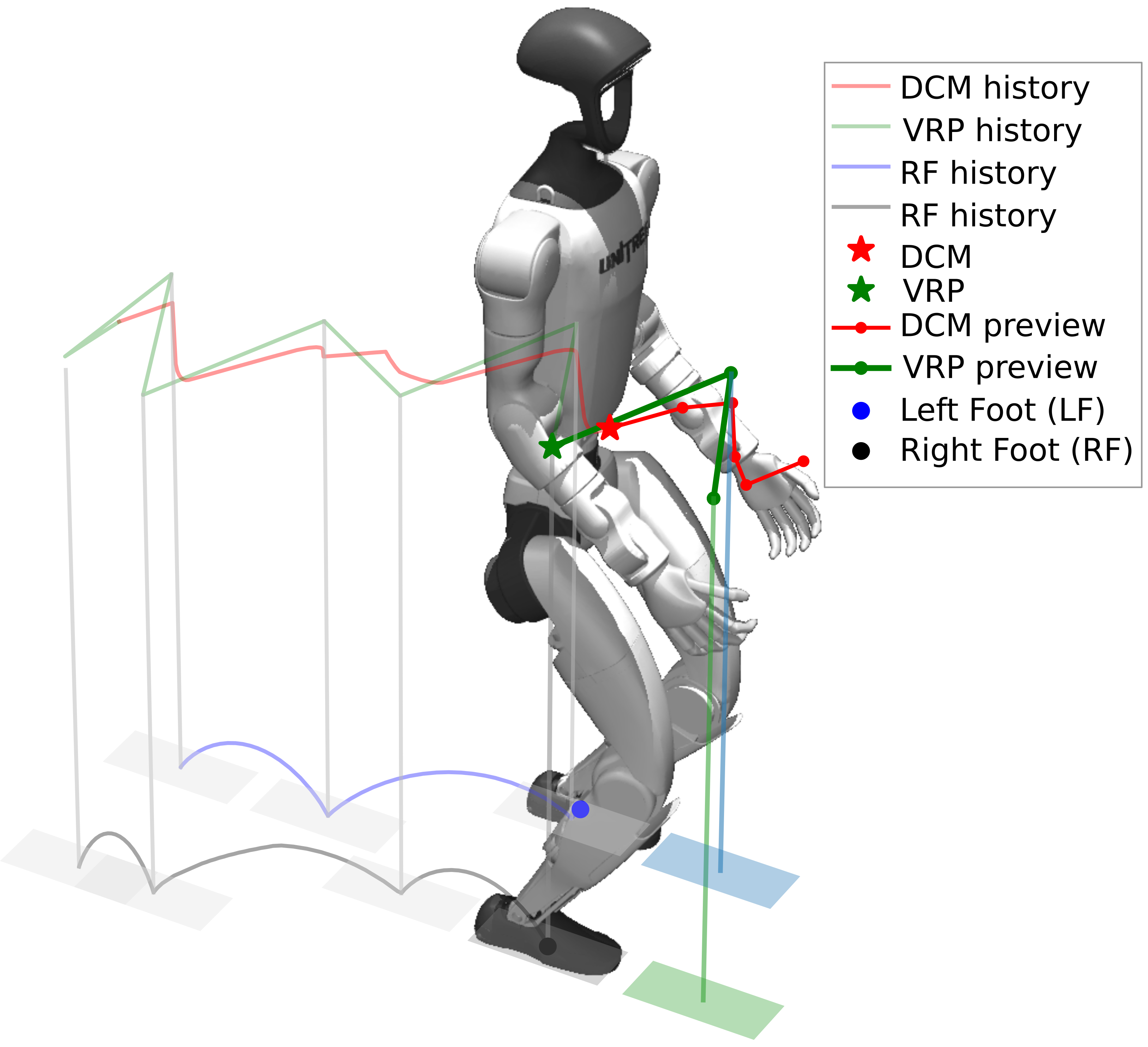}
    \caption{\textbf{DCM and VRP dynamics with footsteps.} 
    The grey (history) and colored (preview) rectangles show the footstep sequence. Bold dots and stars indicate the current foot locations and current DCM/VRP positions, respectively. 
    VRP waypoints ({\color{blue}green}) are derived from these footsteps, and their interpolation defines the VRP trajectory. 
    The DCM trajectory ({\color{red}red}) is then computed from the VRP according to the dynamics in \eqref{eq:dcm_dyn_brief}. 
    Markers indicate the start and end points of each phase. 
    This illustrates how planned footsteps determine VRP positions, which in turn determine the DCM trajectory.}
    \label{fig:dcm_vrp_relation}
\end{figure}

Controlling underactuated humanoid robots requires coordinating a high-dimensional set of degrees of freedom (DoFs) while satisfying complex physical constraints, including feasible ground reaction forces, friction limits, and dynamic stability requirements. A whole-body controller must coordinate full-body joint motions to accomplish diverse tasks while consistently maintaining these constraints. Conventional model-based approaches typically rely on accurate whole-body dynamic models, but their performance can degrade when deployed on real robots due to modeling inaccuracies and the resulting sim-to-real gap. In contrast, end-to-end RL approaches learn control policies directly from data, enhancing robustness through sufficient dynamics randomization during training. However, existing RL methods are often specialized for individual tasks, such as velocity-driven locomotion, specific loco-manipulation scenarios, or retargeting from human motions. The resulting policies often lack generalization to unseen tasks and environments. To enable more generalizable whole-body control for high-DoF humanoid robots, we propose AnSER, an end-to-end learning framework that leverages reduced-order robot dynamics as prior knowledge to improve the generalization and robustness of whole-body control.

Reduced-order robot dynamics have been widely adopted in model-based approaches, where whole-body control is commonly decomposed into high-level trajectory planning and low-level torque optimization. High-level planners typically employ simplified dynamic models, such as the robot CoM or DCM dynamics, instead of full-body robot dynamics. For example, DCM-based walking planners \cite{kajita3DLinearInverted2001, sugiharaRealtimeHumanoidMotion2002, wieberTrajectoryFreeLinear2006, takenakaRealTimeMotion2009, englsbergerThreeDimensionalBipedalWalking2015} generate structured references for CoM/DCM trajectories, footsteps, and timing. Figure \ref{fig:dcm_vrp_relation} illustrates the relation between DCM trajectories and footsteps. These references are subsequently tracked by whole-body controllers \cite{englsbergerTorqueBasedDynamicWalking2018, mesesanDynamicWalkingCompliant2019}, which are typically formulated as weighted quadratic programming (QP)-based task-space controllers. However, this hierarchical decomposition between high-level planners and low-level controllers can limit motion adaptability and feasibility. Since high-level planners often simplify or omit whole-body constraints, such as kinematic limits and self-collision avoidance, they may fail to generate feasible or versatile motions \cite{egleStepTimingAdaptation2023}. For example, simplified footstep planners may restrict stepping strategies to ipsilateral motions, preventing more complex behaviors such as crossover stepping.

Learning-based methods provide an alternative paradigm by allowing policies to learn complex nonlinear mappings directly from data. RL-based approaches \cite{rudin2021learning, ANYmalparkour, zakkaMuJoCoPlayground2025} can incorporate flexible behavioral preferences through reward shaping, where undesirable behaviors such as self-collision can be discouraged by adding corresponding penalty terms. However, despite this flexibility, RL policies remain highly dependent on reward design. Designing effective reward functions is particularly challenging for humanoid robots due to their high-dimensional action spaces and ability to perform diverse whole-body motions. Poorly designed rewards may lead to undesired behaviors, limited task diversity, or policies that fail to generalize beyond the training scenarios.

Recently, hybrid approaches that combine model-based knowledge with RL have shown promising results in humanoid control. For example, RL has been combined with planned footsteps and stepping-stone traversal \cite{duanSimtoRealLearningFootstepConstrained2022a} or reduced-order models (ROMs) to guide policy learning \cite{wangLearning3DBipedal2023,chenReinforcementLearningReducedorder2024}. Other works couple MPC/optimization with RL at the planning layer to augment footstep policies and improve agility/robustness \cite{bangRLaugmentedMPCFramework2024, egleEnhancingModelBasedStep2024}. However, existing methods typically leverage reduced-order models only as external guidance or planning modules, while the resulting whole-body policies remain specialized to individual tasks. This motivates the development of a unified learning framework that integrates model-based prior knowledge to enable more generalizable and robust whole-body control.

Considering the high dimensionality of humanoid robots, we first propose a compact robot state representation that includes only the robot legs and CoM quantities. Unlike conventional state representations that include full-body joint configuration, our representation excludes upper-body joint states while implicitly encoding upper-body motions through their effects on CoM dynamics. Second, we propose an end-to-end learning framework in which the RL policy jointly learns task-space planning and whole-body control. By explicitly learning a task-space planner, our approach produces more interpretable robot behaviors. For example, the policy provides insights into the robot's intended motion by predicting future footstep locations and step timings. Furthermore, we formulate whole-body control as a CoM regulation problem rather than direct full joint-space control. The policy learns to stabilize the robot’s CoM under external disturbances by adjusting only the leg joints, while upper-body joints are excluded from both policy observations and actions. Consequently, upper-body motions are treated as external disturbances to CoM dynamics that the learned controller can compensate for. This formulation enables the upper body to perform arbitrary motions without requiring task-specific training, improving the generality and adaptability of the learned whole-body controller.

Specifically, foot and DCM reference trajectories are generated from the predicted footstep locations and timings to guide policy learning. To improve the utilization of ground reaction forces for balance recovery, we employ an analytically derived ankle-correctable DCM error set to determine when footstep adaptation is necessary. This design prioritizes CoP modulation and triggers step adjustment only when ankle-based stabilization becomes insufficient. For efficient recovery from large disturbances, the policy predicts stepping parameters that drive the measured DCM endpoint to converge quickly to its desired position. We benchmark our method against both a classical model-based DCM planner with whole-body control (WBC) and a purely end-to-end learning baseline, demonstrating improved robustness and faster disturbance recovery.

Our main contributions are summarized as follows:
\begin{itemize}
\item We propose an end-to-end learning framework that jointly learns task-space planning and whole-body control, improving policy interpretability by explicitly predicting future robot behaviors.

\item We introduce \emph{Ankle Recovery Condition} (ARC) during training to encourage exploitation of ground reaction forces for balance recovery, reducing unnecessary stepping and enabling stable locomotion at lower stepping frequencies.

\item We incorporate the Divergent Component of Motion (DCM) endpoint into reward shaping to optimize stepping strategies and achieve efficient recovery from large external disturbances.

\item We design a robot state representation that excludes upper-body joint states while encoding their effects through CoM dynamics, allowing the policy to stabilize arbitrary upper-body motions without task-specific training and enabling efficient generalizable whole-body control.

\item We evaluate our framework on a G1 humanoid robot across diverse unseen scenarios, including terrain disturbances (e.g., soft mats and obstacles) and novel upper-body motions generated through teleoperation and direct human-robot interaction, demonstrating robust generalization.

\end{itemize}


\section{Related Works}
\begin{figure*}
    \centering
    \includegraphics[width=0.9\linewidth]{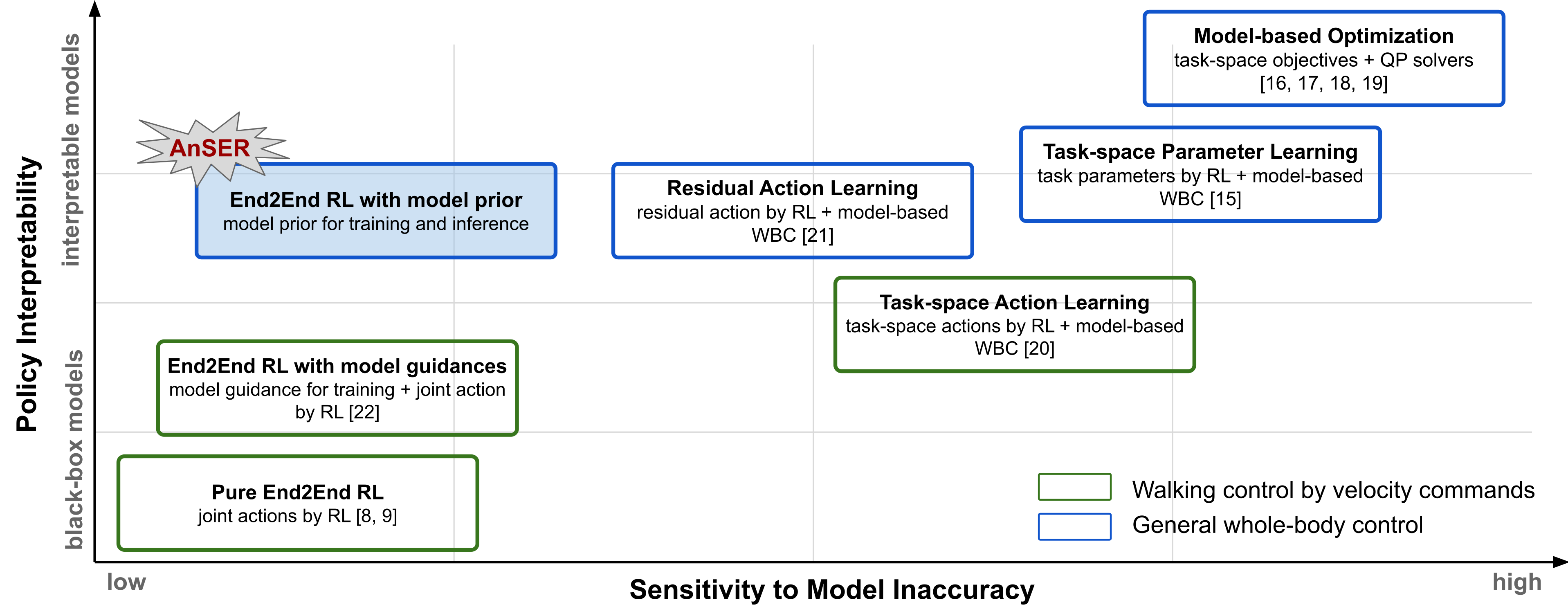}
    \caption{\textbf{Frameworks for Bipedal Locomotion.} Comparison of existing locomotion paradigms in terms of policy interpretability and dependence on accurate robot models. Model-based methods often rely on accurate robot modelling, while learning-based methods decrease the sensitivity to model inaccuracy. However, learned policies are black-box networks, limiting their interpretability, analysis, and reuse in high-level planning and control tasks. Our method, \textit{AnSER}, combines the robustness of end-to-end learning with the interpretability of reduced-order CoM dynamics to learn a generalized whole-body controller. By regulating the robot's CoM, AnSER enables robust locomotion while naturally compensating for arbitrary upper-body motions that are absent from the training data.}
    \label{fig:related_works}
\end{figure*}

Legged robot control has been studied extensively using both model-based and learning-based approaches. Figure \ref{fig:related_works} provides an overview of related works from both domains and positions our method relative to existing approaches in terms of their reliance on robot model knowledge and the interpretability of their control policies.

\subsection{Reduced-order Models for Humanoid Control}
Humanoid robots are generally described by nonlinear, contact-constrained multibody dynamics. Although accurate, these models are high-dimensional and computationally expensive for long-horizon prediction. This motivates models that retain only task-relevant dynamics. For walking, the CoM and its evolution, influenced by the planned contacts and its resulting ground reaction forces is essential. A common approximation is the linear inverted pendulum model (LIPM), which represents the robot by its CoM while assuming a constant CoM height and neglecting centroidal angular-momentum variations \cite{kajita3DLinearInverted2001}. Its second-order CoM dynamics are frequently propagated within model predictive control schemes \cite{diedamOnlineWalkingGait2008, romualdiOnlineNonlinearCentroidal2022a}. However, constrained MPC typically propagates these dynamics through discretized recursions and repeatedly solves an optimization problem over the prediction horizon. An alternative method is to utilize the DCM, which exactly decomposes the second-order CoM dynamics into two first-order systems: unstable DCM dynamics and stable CoM dynamics \cite{takenakaRealTimeMotion2009,englsbergerThreeDimensionalBipedalWalking2015}. Thus, the DCM does not introduce a further approximation of the translational CoM dynamics; it is reduced because angular dynamics are not represented. For piecewise-defined CoP or VRP trajectories, both systems admit analytical propagation \cite{kamiokaSimultaneousOptimizationZMP2018,mesesanConvexPropertiesCenterofMass2018}. As only the DCM dynamics are unstable, it is sufficient to stabilize the DCM, while the CoM follows naturally. This compact analytical formulation avoids the need for iterative numerical integration during prediction and lends itself well to massively parallel GPU computation.

\subsection{Reinforcement Learning for Humanoid Control}
Recent advances in GPU-accelerated physics simulators, such as IsaacSim/IsaacGym, MuJoCo-MJX, and Genesis, have significantly alleviated the sample efficiency bottleneck of RL. By leveraging massively parallel simulation, RL-based methods have achieved remarkable success in robotic control, particularly for high-DoF legged and humanoid robots. Rudin et al.~\cite{rudin2021learning} demonstrated that robust quadrupedal locomotion policies can be learned within minutes using massively parallel training in IsaacGym. Extending beyond periodic locomotion on structured terrain, Hoeller et al.~\cite{ANYmalparkour} showed that end-to-end RL can synthesize agile parkour behaviors over highly challenging environments, highlighting the potential of learning-based methods for generating complex, dynamic motions.

Despite these impressive results, learning control policies from scratch often requires carefully engineered reward functions, making reward design a labor-intensive and task-specific process. To reduce this dependence, several studies have integrated RL with model-based control. Rather than directly learning low-level control commands, Duan et al.~\cite{9561705} trained an RL policy to predict high-level foothold setpoints, which are then executed by a model-based planner and low-level controller. Similarly, Egle et al.~\cite{egleEnhancingModelBasedStep2024} learned task-space planning variables, including feasible foothold regions and step timing, to improve the disturbance rejection capability of a model-based controller under external pushes. By restricting RL to high-level decision variables, these hybrid approaches reduce the learning complexity and improve sample efficiency. However, their performance remains strongly dependent on the accuracy of the underlying system model and the reliability of the model-based controller.

To improve robustness against model inaccuracies, Yan et al.~\cite{11395590} proposed learning residual control actions that compensate for unmodeled dynamics, including errors in friction, damping, and inertial parameters. From a different perspective, Li~\cite{11346966} introduced a structured reward shaping framework based on control Lyapunov functions to guide policy optimization. Instead of manually designing numerous reward terms, their approach constructs rewards by tracking continuous task-space objectives, such as the CoM and foot trajectories, providing a more systematic and principled framework for reward design.

\subsection{Motion Retargeting for Humanoid Control}
Unlike conventional RL-based locomotion methods, which primarily focus on controlling the lower-body joints for locomotion, motion retargeting enables humanoids to perform versatile whole-body behaviors by imitating human motions. In this paradigm, humanoid control is formulated as a whole-body motion tracking problem. A typical pipeline consists of two stages. First, human motions are retargeted to the robot morphology using either inverse-kinematics-based optimization~\cite{araujo2025retargetingmattersgeneralmotion, yang2025omniretarget} or learning-based retargeting methods~\cite{ImitationNet, 11397669}. The resulting robot reference motions are then used to train RL policies that reproduce the desired behaviors in physics simulators.

To improve the scalability of whole-body motion learning, Yan et al.~\cite{ictrl} constrained the policy action space to a reference-centered subspace and introduced an automatic curriculum based on motion difficulty, significantly improving sample efficiency over large-scale motion datasets. Ze et al.~\cite{ze2025twist, ze2025twist2scalableportableholistic} adopted a teacher--student learning framework to improve policy robustness during sim-to-real transfer, enabling reliable whole-body teleoperation using human motion capture systems and virtual reality devices. Beyond tracking prerecorded motions, Liao et al.~\cite{liao2025beyondmimicmotiontrackingversatile} incorporated a latent diffusion model to synthesize diverse, previously unseen human-like motions, substantially expanding the executable behaviors. Furthermore, Gabriel et al.~\cite{margolis2025softmimiclearningcompliantwholebody} introduced joint stiffness in the policy input, allowing the controller to adapt its tracking behavior according to user-specified compliance and thereby achieve more compliant whole-body motions.

Despite their impressive performance, existing motion retargeting methods share two important limitations. First, they require full-body human kinematic observations as input, restricting the control interface to motion capture systems or other full-body sensing modalities. Second, their objective is to accurately reproduce reference kinematics, with limited consideration of robot dynamics. As a result, the learned behaviors are not explicitly optimized for dynamic balance, disturbance rejection, or force generation during interaction and manipulation. Consequently, when subjected to external disturbances or physically demanding tasks, strict kinematic tracking may conflict with the robot's dynamic feasibility and stability requirements.

In contrast, our method aims to learn a general-purpose whole-body controller by incorporating robot CoM dynamics as prior knowledge, enabling the policy to generalize to unseen motions without task-specific training. By introducing footstep parameters as additional policy outputs, our framework explicitly optimizes disturbance recovery strategies while improving policy interpretability by providing insights into the robot's future behaviors.

\section{Preliminaries}
\subsection{Divergent Component of Motion (DCM)}
\label{sec:prel_dcm_brief}
The DCM is a variable that exposes the unstable part of the CoM dynamics and has become a central concept in humanoid locomotion planning \cite{takenakaRealTimeMotion2009,englsbergerThreeDimensionalBipedalWalking2015}. For a nominal CoM height $\Delta z$ over flat terrain, we define
\begin{equation}
    \boldsymbol{\xi} = \boldsymbol{x} + b\,\dot{\boldsymbol{x}},
\end{equation}
where $\boldsymbol{x}, \boldsymbol{\dot{x}}$ are the CoM position and velocity, and time constant $b = \sqrt{\Delta z / g}$. Introducing the Virtual Repellent Point (VRP) $\boldsymbol{v}$, the DCM follows the first-order dynamics
\begin{equation}
    \dot{\boldsymbol{\xi}} = \frac{1}{b}\bigl(\boldsymbol{\xi} - \boldsymbol{v}\bigr).
    \label{eq:dcm_dyn_brief}
\end{equation}
Here, $\boldsymbol{v}$ encodes the total force $\boldsymbol{f}$ acting on the CoM via $\boldsymbol{f}=\frac{m}{b^2}(\boldsymbol{x} - \boldsymbol{v})$, where $m$ is the total mass. On level ground and with constant height, the horizontal DCM coincides with the (instantaneous) capture point \cite{prattCapturePointStep2006}.
\subsection{DCM Trajectory Generation}
\label{sec:prel_dcm_traj}

The VRP encodes the total force acting on the CoM. With constant centroidal angular momentum, its vertical projection must remain within the support polygon to guarantee feasible contact forces. Figure~\ref{fig:dcm_vrp_relation} illustrates the relation between footsteps and VRP, while Fig.~\ref{fig:ankle_set} provides an overview of the support region and the corresponding VRP waypoints $\boldsymbol{v}_i$ under different support conditions. 
For a single phase of duration $T$, with VRP endpoints $\boldsymbol{v}_0$ and $\boldsymbol{v}_T$, the DCM admits a closed-form solution.  
The \emph{backward (terminal-value) solution} computes the initial DCM from a desired terminal value:
\begin{equation}
    \boldsymbol{\xi}(0) = \alpha(T)\,\boldsymbol{v}_0 + \beta(T)\,\boldsymbol{v}_T + \gamma(T)\,\boldsymbol{\xi}_T,
\end{equation}
where $\alpha,\beta,\gamma$ are scalar functions of $T$.  
Similarly, the \emph{forward (initial-value) solution} propagates the measured state $\boldsymbol{\xi}_0$ to obtain the phase endpoint $\boldsymbol{\xi}_T$.  
The forward and backward solutions thus represent two alternative approaches: the backward solution enforces convergence to a desired terminal state for planned footsteps, while the forward solution guarantees the consistency of the current state. Over multiple phases, trajectories are obtained by enforcing equality constraints between consecutive phase endpoints, ensuring continuity across the full preview horizon \cite{englsbergerThreeDimensionalBipedalWalking2015, mesesanConvexPropertiesCenterofMass2018}.

\subsection{DCM error separation / Ankle strategy}
\label{sec:ankle_strategy}

The ankle strategy corresponds to modulating the commanded VRP in the horizontal plane within the boundaries of the current support polygon. This allows the robot to reject small disturbances by shifting the CoP without altering the pre-planned footstep sequence. Following \cite{mesesanOnlineDCMTrajectory2021}, the measured DCM error $\tilde{\boldsymbol{\xi}}(t) = \boldsymbol{\xi}(t) - \boldsymbol{\xi}_{\mathrm{ref}}(t)$
can be decomposed into two parts,
\begin{equation}
    \tilde{\boldsymbol{\xi}}(t) = \tilde{\boldsymbol{\xi}}_{\mathrm{ankle}}(t) + \tilde{\boldsymbol{\xi}}_{\mathrm{step}}(t),
\end{equation}
where $\tilde{\boldsymbol{\xi}}_{\mathrm{ankle}}$ is the component that can be corrected using ankle modulation, and $\tilde{\boldsymbol{\xi}}_{\mathrm{step}}$ is the remaining component that must be corrected through a stepping strategy.  

In single support, we denote $\tilde{V}$ as the set of feasible VRP adjustments that remain within the rectangular support polygon of half-length $l$ and half-width $w$:
\begin{equation}
    \tilde{V} = \left\{ \tilde{\boldsymbol{v}} = (x \ y \ 0)^{\top} \,\big|\, -l \leq x \leq l,\; -w \leq y \leq w \right\}.
\end{equation}
The set of VRP adjustments available to the ankle strategy for the $j$-th footstep, expressed in world coordinates, is
\begin{equation}
    \tilde{V}_{\mathrm{ankle}} = \{\boldsymbol{R}_{z,j} \tilde{\boldsymbol{v}} \mid \tilde{\boldsymbol{v}} \in \tilde{V}\},
\end{equation}
where $\boldsymbol{R}_{z,j} \in SO(3)$ is the rotation matrix associated with the foot yaw orientation.

The corresponding set of correctable DCM errors until the end of the current step, denoted $\tilde{\Xi}_{\mathrm{ankle}}$, is obtained from the recursive DCM solution. For the current and next transition phases, the DCM dynamics can be written as
\begin{equation}
\begin{aligned}
    \tilde{\boldsymbol{\xi}}_{0,1} &= \alpha_{1}\,\tilde{\boldsymbol{v}}_{0,1} + \beta_{1}\,\tilde{\boldsymbol{v}}_{T,1} + \gamma_{1}\,\tilde{\boldsymbol{\xi}}_{T,1},\\
    \tilde{\boldsymbol{\xi}}_{0,2} &= \alpha_{2}\,\tilde{\boldsymbol{v}}_{0,2} + \beta_{2}\,\tilde{\boldsymbol{v}}_{T,2} + \gamma_{2}\,\tilde{\boldsymbol{\xi}}_{T,2}.
\end{aligned}
\end{equation}
Assuming that the DCM error vanishes at the end of the next phase ($\tilde{\boldsymbol{\xi}}_{T,2} = 0$), and using the continuity conditions $\tilde{\boldsymbol{\xi}}_{T,1}=\tilde{\boldsymbol{\xi}}_{0,2}$ and $\tilde{\boldsymbol{v}}_{T,1} = \tilde{\boldsymbol{v}}_{0,2}$, the set of DCM errors that can be corrected by the ankle strategy becomes
\begin{equation}
    \tilde{\Xi}_{\mathrm{ankle}} = \alpha_{1}\,\tilde{V}_0 + \left(\beta_{1} + \gamma_{1}\alpha_{2}\right)\tilde{V}_T.
\end{equation}

\begin{figure}
    \centering
    \vspace{5pt}
    \includegraphics[width=0.99\linewidth]{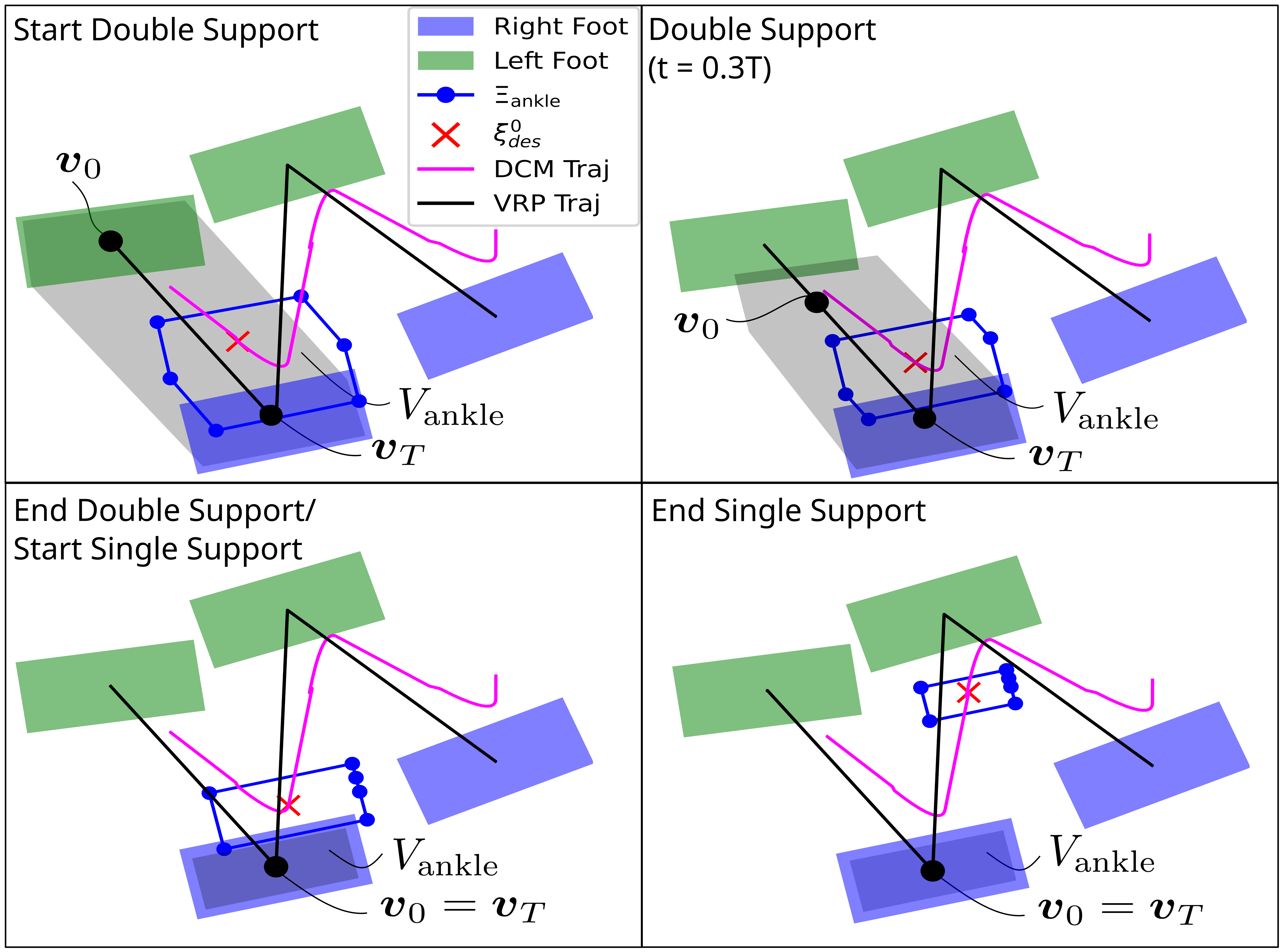}
    \caption{\textbf{Ankle Region Evolution for different Support Conditions.} The Virtual Repellent Point (VRP) represents the total force acting on the center of mass (CoM). Its feasible adjustments are bounded by the support polygon of the current contact configuration, from which the correctable DCM error set $\Xi_{\mathrm{ankle}}$ is computed (Sec.~\ref{sec:ankle_strategy}).}
    \label{fig:ankle_set}
\end{figure}

During single support, $\tilde{V}_0$ and $\tilde{V}_T$ coincide. 
In double support, however, $\tilde{V}_0$ corresponds to the entire support polygon formed by both stance feet, while $\tilde{V}_T$ is restricted to the support region of the upcoming single-support foot. Different support conditions with the corresponding sets $\tilde{\Xi}_{\mathrm{ankle}}$ and $\tilde{V}_{\mathrm{ankle}}$ are illustrated in Fig.~\ref{fig:ankle_set}.

\begin{figure*}[]
    \centering
    \vspace{5pt}
    \includegraphics[width=0.98\textwidth]{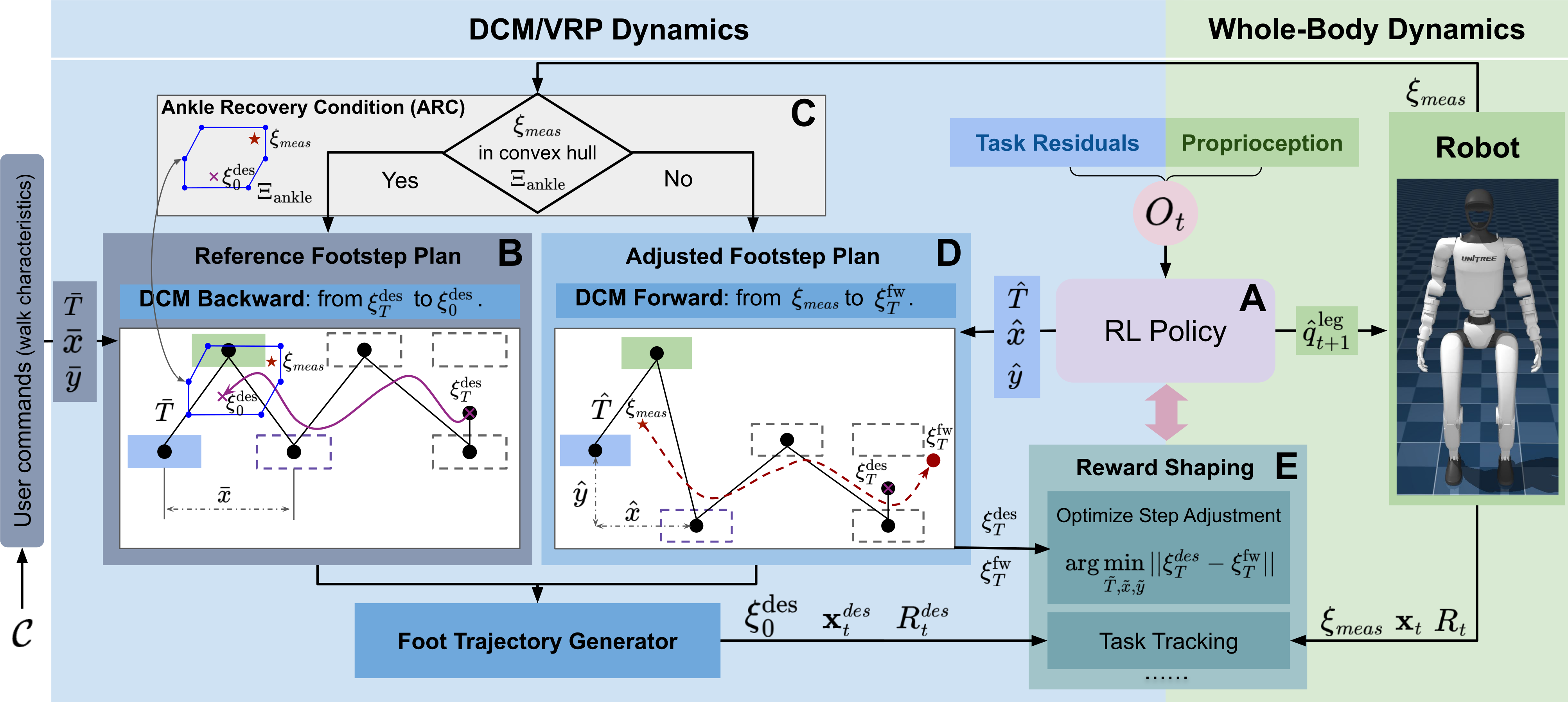}
    \caption{\textbf{Framework overview.} In our framework, the RL policy simultaneously learns whole-body control (\textcolor{darkgreen}{green}) and task-space planning ({\color{blue}blue}). As input, the policy receives proprioceptive states and task residuals (e.g., tracking errors), and outputs target joint positions $\hat{\boldsymbol{q}}_{t+1}$ for execution. 
    In parallel, it predicts the next footstep location and timing $(\hat{x}, \hat{y}, \hat{T})$. 
    The Ankle Recovery Condition (ARC) determines whether the nominal reference plan $(\bar{x}, \bar{y}, \bar{T})$, derived from the user velocity command $(v_{\mathrm{ref}}, \omega_{\mathrm{ref}})$ and a backward DCM preview, is sufficient, or whether the policy’s adapted step parameters should be used. 
    In the latter case, a forward DCM computation ensures consistency with the measured state $\boldsymbol{\xi}_{\mathrm{meas}}$. 
    Reward shaping combines task tracking with penalties on terminal DCM deviation, enabling the policy to learn adaptive stepping for disturbance recovery.}
    \label{fig:overview}
\end{figure*}
\section{Methodology}
Our framework unifies whole-body control and task-space planning within a single RL policy. 
For enhanced balancing, the central idea is to favor ankle-based recovery through CoP modulation whenever possible, and to trigger adaptive stepping only when necessary. 
Figure~\ref{fig:overview} summarizes the full pipeline.
The control loop consists of:
\begin{itemize}
    \item \textbf{RL policy (Sec.~\ref{sec:plan_control})} computes joint targets and candidate footstep parameters.
    \item \textbf{Reference footstep planning (Sec.~\ref{sec:nominal})} generates a reference plan from velocity commands via backward DCM computation.
    \item \textbf{Ankle Recovery Condition (Sec.~\ref{sec:adapt_cond})} determines whether CoP modulation is sufficient or whether additional footstep adjustment is needed.
    \item \textbf{Adjusted footstep planning (Sec.~\ref{sec:fwd_replan})} recomputes the DCM trajectory if adaptation is triggered.
    \item \textbf{Reward design (Sec.~\ref{sec:reward})} shapes learning by combining task tracking with penalties on DCM error.
    \item \textbf{Dynamics randomization (Sec.~\ref{sec:DR})} gives the details of randomization on robot dynamics.
\end{itemize}
The subsections below detail each stage.

\subsection{RL Policy}
\label{sec:plan_control}
In our framework, the RL policy jointly learns high-level motion planning in task space and low-level whole-body control in joint space. At the high level, the policy adapts stepping to recover effectively from disturbances, thereby improving locomotion robustness. At the low level, it generates joint position commands for legs to accurately track the planned task-space trajectories. To achieve this, the policy receives observations that combine both proprioceptive data and task-space residuals.

The proprioceptive data include leg-joint positions $\mathbf{q}_t^{\text{leg}}$ and the previous RL action $\mathbf{a}_{t-1}$. Our policy input also includes task-level residuals $\mathcal{E}$, consisting of
\begin{equation}
\begin{aligned}
\mathbf{e}_{\text{com}}^{\text{r}} &= \mathbf{R}^{\text{des}}_{\text{com}} \circleddash \mathbf{R}_{\text{com}}, \\
\mathbf{e}_{\text{com}}^{\text{p}} &= \mathbf{x}^{\text{des}}_{\text{com}} - \mathbf{x}_{\text{com}}, \\
\mathbf{e}_{\text{ft}}^{\text{r}} &= \mathbf{R}^{\text{des}}_{\text{ft}} \circleddash \mathbf{R}_{\text{ft}}, \\
\mathbf{e}_{\text{ft}}^{\text{p}} &= \mathbf{x}^{\text{des}}_{\text{ft}} - \mathbf{x}_{\text{ft}} ,
\end{aligned}
\end{equation}
where $\mathbf{e}_{\text{com}}^{\text{r}}, \mathbf{e}_{\text{com}}^{\text{p}} \in \mathbb{R}^3$ and $\mathbf{e}_{\text{ft}}^{\text{r}}, \mathbf{e}_{\text{ft}}^{\text{p}} \in \mathbb{R}^6$ are the error vectors of orientation and position of CoM and two feet, respectvely. The operator $\circleddash$ computes the difference between rotation matrices, expressed in Euler angles.

In addition, our policy supports an extended set of user-defined commands, denoted by $\mathcal{C}$, to regulate various walking characteristics beyond locomotion velocity alone:
\begin{equation}
\mathcal{C} = \left[ v_{xy}, \omega_{z}, r_{y}^{\mathrm{com}}, h_{z}^{\mathrm{com}}, h_{z}^{\mathrm{step}}, w_{\mathrm{step}}, f_{\mathrm{step}} \right].
\label{eq:cmd}
\end{equation}

Here, $v_{xy}$ and $\omega_{z}$ specify the desired linear and angular velocities, respectively. The parameters $r_{y}^{\mathrm{com}}$ and $h_{z}^{\mathrm{com}}$ define the desired torso pitch angle and CoM height. In addition, the stepping characteristics can be adjusted through the desired step height $h_{z}^{\mathrm{step}}$, step width $w_{\mathrm{step}}$, and step frequency $f_{\mathrm{step}}$. This extended command space enables users to directly modify both locomotion behavior and gait characteristics during execution.

Overall, the policy input at time step $t$ consists of a history of observations, $\left\{ \mathbf{o}_{t-k} \ldots \mathbf{o}_{t} \right\}$, where each observation is defined as
\begin{equation}
    \mathbf{o}_t = \left[ \mathcal{C}, \mathbf{q}_t^{\mathrm{leg}}, \mathbf{a}_{t-1}, \mathcal{E}_t \right]
\end{equation}

At each control cycle, the policy receives proprioceptive states and task residuals as input, and outputs
\begin{equation}
    \mathbf{a}_t = \left[ \hat{x}_t,\quad \hat{y}_t,\quad \hat{T}_t,\quad \hat{\boldsymbol{q}}_{t+1}^{\mathrm{leg}} \right],
\end{equation}
where $(\hat{x}_t,\hat{y}_t,\hat{T}_t)$ denote the next footstep location and timing, and $\hat{\boldsymbol{q}}_{t+1}^{\mathrm{leg}}$ are target leg-joint positions tracked by a low-level PD controller. Joint commands are always executed, while the footstep parameters are applied only when the Ankle Recovery Condition requires them.

\subsection{Nominal plan and backward DCM}\label{sec:nominal}
From the user’s command $(v_{\mathrm{xy}},\omega_{\mathrm{z}}, w_{\mathrm{step}}, f_{\mathrm{step}})$, a nominal footstep sequence is generated. Given this nominal plan and a desired terminal DCM $\boldsymbol{\xi}^{\mathrm{des}}_T$, we compute the reference DCM trajectory by solving the DCM dynamics backward in time:
\begin{equation}
    \boldsymbol{\xi}(t) = \alpha(t)\,\boldsymbol{v}_0
                        + \beta(t)\,\boldsymbol{v}_T
                        + \gamma(t)\,\boldsymbol{\xi}^{\mathrm{des}}_T,
    \qquad t\in[0,T],
    \label{eq:bwd_dcm_rl}
\end{equation}
where $\alpha,\beta,\gamma$ are scalar coefficients depending on the phase duration $T$ (see Sec.~\ref{sec:prel_dcm_brief}).  
This recursion produces an initial DCM waypoint $\boldsymbol{\xi}_0^{\mathrm{des}}$ that is consistent with the terminal objective. From this, we compute the set of correctable DCM errors as:
\begin{equation}
    \Xi_{\mathrm{ankle}} = \boldsymbol{\xi}_0^{\mathrm{des}} + \tilde{\Xi}_{\mathrm{ankle}},
\end{equation}
which will be used in Sec. \ref{sec:adapt_cond}.

\subsection{Ankle Recovery Condition}\label{sec:adapt_cond}
The goal is to force the RL policy to use CoP pressure modulation whenever possible and avoid excessive stepping. To achieve this, we use the set of correctable DCM errors with the ankle strategy $\Xi_{\mathrm{ankle}}$ as the decision criterion for whether to follow the reference step plan provided by the user input or to switch to the policy output that adapts the step plan for disturbance recovery. At runtime, the measured DCM $\boldsymbol{\xi}_{\mathrm{meas}}$ is compared to the reference set of feasible corrections $\Xi_{\mathrm{ankle}}$. If
\begin{equation}
    \boldsymbol{\xi}_{\mathrm{meas}} \in \Xi_{\mathrm{ankle}},
\end{equation}
Then the nominal plan is kept, and no step adaptation is applied. Otherwise, a new step is generated from the policy output $(\hat{x},\hat{y},\hat{T})$.

\subsection{Step adaptation and forward replanning}\label{sec:fwd_replan}
For an adapted step, the DCM trajectory is recomputed forward from the measured state:
\begin{equation}
    \boldsymbol{\xi}(t) = \alpha(t)\,\hat{\boldsymbol{v}}_0
                        + \beta(t)\,\hat{\boldsymbol{v}}_T
                        + \gamma(t)\,\boldsymbol{\xi}_{\mathrm{meas}},
    \qquad t\in[0,\hat{T}],
    \label{eq:fwd_dcm_rl}
\end{equation}
with $\hat{\boldsymbol{v}}_0,\hat{\boldsymbol{v}}_T$ defined from the adapted footstep $(\hat{x},\hat{y},\hat{T})$.  
This guarantees consistency with the current measured state, while producing a new terminal value $\boldsymbol{\xi}^{\mathrm{fw}}_T$. Since $\hat{\boldsymbol{v}}_0$ and $\hat{\boldsymbol{v}}_T$ depend on the adapted footstep $(\hat{x},\hat{y},\hat{T})$, the placement and timing directly determine $\boldsymbol{\xi}^{\mathrm{fw}}_{T}$ via \eqref{eq:fwd_dcm_rl}.
Thus, during training, the policy is rewarded according to  
\begin{equation}
    r = \big\lVert \boldsymbol{\xi}_{T}^{\mathrm{fw}} - \boldsymbol{\xi}_{T}^{\mathrm{des}} \big\rVert _2,
\label{eq:dcm_terminal}
\end{equation}
which encourages footstep adjustments that minimize the deviation between the forward-predicted terminal DCM $\boldsymbol{\xi}_{T}^{\mathrm{fw}}$ and the desired target $\boldsymbol{\xi}_{T}^{\mathrm{des}}$.


\subsection{Reward Design}
\label{sec:reward}

\begin{table}[t]
\vspace{5pt}
\caption{\textbf{Reward functions.}}
\centering
\renewcommand{\arraystretch}{1.35}
\begin{tabular}{clcccc}
\toprule
 & Reward & Expression & Distance $D$ & $\omega$ & $\lambda$ \\
\midrule

\multirow{4}{*}{\rotatebox{90}{\textbf{Ankle/Step}}}
 & $\mathcal{R}^{step}$ & $\delta_{\text{step}} \cdot \omega e^{-\lambda D}$ & $\|\xi_T^{\text{fw}} - \xi_T^{des}\|_2$ & 20 & 5 \\
 & $\mathcal{R}^{ank}$  & $-\omega \cdot \delta_{\text{step}}$ & -- & 2 & -- \\
 & \multicolumn{5}{l}{$\displaystyle
\delta_{\text{step}} =
 \begin{cases}
 0 & \xi_{\text{meas}} \in \Xi_{\mathrm{ankle}} \ \\
 1 & \text{otherwise}
 \end{cases}$} \\

\midrule
\multirow{5}{*}{\rotatebox{90}{\textbf{Task Tracking}}}
 & $\mathcal{R}^{base}$  & \multirow{4}{*}{$\omega e^{-\lambda D}$} & $\|R^{des}_{base} \circleddash R_{base}\|_2$ & 15 & 20 \\[2pt]
 & $\mathcal{R}^{dcm}$   &  & $\|\xi^{0}_{des} - \xi_{meas}\|_2$ & 30 & 20 \\[2pt]
 & $\mathcal{R}^{ft\_r}$ &  & $\|R^{des}_{ft} \circleddash R_{ft}\|_2$ & 20 & 20 \\[2pt]
 & $\mathcal{R}^{ft\_p}$ &  & $\|\mathbf{x}^{des}_{ft} - \mathbf{x}_{ft}\|_2$ & 25 & 20 \\[2pt]

\midrule
\multirow{6}{*}{\rotatebox{90}{\textbf{Regularization}}}
 & $\mathcal{R}^{acc}$  & $e^{-\lambda D}$ & $\|\dot{q}_t - \dot{q}_{t-1}\|_2$ & 0.1 & 0.01 \\
 & $\mathcal{R}^{act}$  & $e^{-\lambda D}$ & $\|a_t - a_{t-1}\|_2$ & 0.1 & 1.0 \\
 & $\mathcal{R}^{con}$  & \multicolumn{4}{l}{$\begin{cases}$-20$ & \text{if anomaly contacts occur} \\ $0$ & \text{otherwise} \end{cases}$}\\[10pt]
 & $\mathcal{R}^{term}$ & \multicolumn{4}{l}{$\begin{cases}$-100$ & \text{if early termination}\\ $0$ & \text{otherwise}  \end{cases}$} \\

\bottomrule
\end{tabular}
\label{tab:rewards}
\end{table}

Our reward formulation comprises three components: ankle and stepping rewards, task-space tracking rewards, and regularization terms (see Table \ref{tab:rewards}).

To optimize the recovery strategy of the RL policy, we introduce two reward functions: $\mathcal{R}^{ank}$ and $\mathcal{R}^{step}$. The ankle reward $\mathcal{R}^{ank}$ penalizes cases where $\boldsymbol{\xi}_{\mathrm{meas}}$ lies outside the convex hull $\Xi_{ankle}$, forcing the robot to exploit ground reaction forces whenever possible for push recovery. The stepping reward $\mathcal{R}^{step}$ guides the policy to find stepping parameters, i.e., foot placement and step timing, for rapid recovery to the reference plan. To achieve this, $\mathcal{R}^{step}$ optimizes the DCM end-point $\xi_{T}^{\mathrm{fw}}$ to close to the desired end-point $\xi_{T}^{\mathrm{des}}$. Overall, our policy enhances the push recovery by exploiting ground reaction forces to avoid unnecessary steps and optimizing stepping parameters to retain balance rapidly.

The task rewards encourage the robot to track reference trajectories that are generated by DCM and foot planners, following the user's high-level commands $\mathcal{C}$. We apply different regularization rewards on the policy action $a_t$, joint velocity $\dot{q}_t$ to smooth the robot motion. Sparse penalties are used to avoid early termination and self-collisions.

\begin{table}[]
\vspace*{5pt}
\centering
\resizebox{0.95\linewidth}{!}{%
\begin{tabular}{lccc}
Parameter & Unit & Distribution & Operator \\ \midrule
\multicolumn{4}{c}{\textbf{Sensory Noise}}  \\ \midrule
Linear velocity & m/s  & $\mathcal{N}(0, 0.1)$ & additive \\
Angular velocity & rad/s  & $\mathcal{N}(0, 0.1)$ & additive \\
Joint position & rad  & $\mathcal{N}(0, 0.005)$ & additive \\
Joint velocity & rad/s & $\mathcal{N}(0, 0.1)$ & additive \\ \midrule
\multicolumn{4}{c}{\textbf{Dynamics Uncertainty}}  \\ \midrule
Body mass & - & $\mathcal{U}(0.9, 1.3)$ & scaling \\
Joint friction & - & $\mathcal{U}(0.5, 1.5)$ & scaling  \\
Joint damping & - & $\mathcal{U}(0.5, 1.5)$ & scaling \\ 
Stiffness & - & $\mathcal{U}(0.8, 1.2)$ & scaling \\ 
Damping & - & $\mathcal{U}(0.5, 1.5)$ & scaling \\ \midrule
\multicolumn{4}{c}{\textbf{Terrain}}  \\ \midrule
Floor friction & - & $\mathcal{U}(0.4, 1.0)$ & -  \\
\end{tabular}
}
\caption{\textbf{Randomization} are applied for real-world uncertainties. }
\label{tab:domain_randomization}
\end{table}
\subsection{Dynamics Randomization} 
\label{sec:DR}
During training, we apply randomization to different sources to capture the variability in real-world environments. The details are illustrated in Table \ref{tab:domain_randomization}. The same randomization is applied in the learning-based baseline method, too.

\section{Simulation Results}
In this section, we first describe the implementation details in Sec.~\ref{sec:implementation}, followed by the baseline methods used for comparison in Sec.~\ref{sec:baseline}. We then present quantitative and qualitative evaluations in Secs.~\ref{sec:quantitative} and \ref{sec:qualitative}, based on the metrics introduced in Sec.~\ref{sec:metric}.

\subsection{Implementation Details}
\label{sec:implementation}
The RL policy is implemented as a multi-layer perceptron (MLP) consisting of 8 fully connected layers, each with 256 hidden units, followed by an Exponential Linear Unit (ELU) \cite{fernandez2024stable} activation function.  

Our training framework is implemented in JAX, with the physics simulator MJX~\cite{todorov2012mujoco} enabling GPU-based parallelization. Training uses 4096 parallel environments, each with an episode length of 0.4~s. The simulator runs at 500~Hz, while the policy outputs control and planning commands at 250~Hz. Random pushes are applied to the robot in the $xy$-plane, with push force sampled uniformly from $[-200, 200]$~N in random directions. The push duration is sampled between 0.1 and 0.2 seconds.  

We evaluate our method on two bipedal robots: Kangaroo~\cite{roigHardwareDesignControl2022} and Unitree G1. Both robots have two fully actuated 6-DoF legs with flat feet. 

\begin{table*}[]
\centering
\caption{\textbf{Comparison on Walking in Simulations.} Different methods are compared on defined metrics for bipedal walking w/w.o. external disturbances. Bold indicates the best performance. Note that no randomization is applied to the model-based baseline.}
\resizebox{\textwidth}{!}{%
\begin{tabular}{lcccccccccccc}
& \multicolumn{6}{c}{\textbf{Without disturbances}} & \multicolumn{6}{c}{\textbf{With disturbances}} \\ \cmidrule(lr){2-7} \cmidrule(lr){8-13}
 & \multicolumn{2}{c}{GoD (m)} &
  \multicolumn{2}{c}{CoT} & 
  \multicolumn{2}{c}{DCM Track (cm)} & 
  \multicolumn{2}{c}{Recover time (s)} & 
  \multicolumn{2}{c}{Trajectory deviation (cm)} & 
  \multicolumn{2}{c}{SR (\%)} \\ \cmidrule(lr){2-3} \cmidrule(lr){4-5} \cmidrule(lr){6-7} \cmidrule(lr){8-9} \cmidrule(lr){10-11} \cmidrule(lr){12-13}
            & Kangaroo & G1 & Kangaroo & G1 & Kangaroo & G1 & Kangaroo & G1 & Kangaroo & G1 & Kangaroo & G1 \\ \midrule
Playground \cite{zakkaMuJoCoPlayground2025}  & 2.21    & 0.71  & 5.03     &  2.17  & -        & -  & 0.54     & 0.49  & 19.91    &  10.19  & 99.50    & 94.87  \\
model-based \cite{egleStepTimingAdaptation2023} & \textbf{0.149}     & \textbf{0.06}  & \textbf{0.57}     & 0.91 & \textbf{0.284}    & \textbf{0.59}  & 0.27     & 0.51  & 21.19    &  18.82  & 87.43    & 80.40  \\ \midrule
AnSER       & 0.38     & 0.17  & 0.91     &  \textbf{0.64}  & 0.38 & 0.65 & \textbf{0.22} & \textbf{0.34} & \textbf{12.31}    &  \textbf{6.64}  & \textbf{99.90}    & \textbf{100}
\end{tabular}%
}
\label{tab:quantitative}
\end{table*}

\subsection{Baseline Methods}
\label{sec:baseline}
To assess the effectiveness of our approach for locomotion and push recovery, we compare against both end-to-end learning and model-based baselines.

\subsubsection{End-to-End RL Baseline} 
We adopt Mujoco Playground~\cite{zakkaMuJoCoPlayground2025}, an open-source MJX-based framework, as our end-to-end learning baseline. Playground achieves robust locomotion by incorporating numerous heuristics that encourage specific walking characteristics, such as velocity tracking, foot swing height, foot air time, and foot slippage. However, these heuristics do not fully correspond to natural walking behaviour; for instance, when switching the support foot, robots generate natural y-axis velocities regardless of the commanded velocities. Moreover, purely end-to-end learning approaches like Playground may converge to sub-optimal solutions, such as consistently taking rapid steps solely to maintain balance.  

In contrast, our approach explicitly incorporates the analytical robot model, i.e., DCM/VRP dynamics, to improve the learned behaviour. By additionally integrating the ankle and stepping strategies, our method enhances ground reaction force exploitation for balancing, reducing unnecessary steps. Meanwhile, our policy optimizes foot placement and step timing to recover from large disturbances efficiently.

\subsubsection{Model-Based Locomotion Baseline}
Model-based methods for bipedal walking and push recovery typically rely on optimization-based whole-body controllers (WBCs) that compute joint torques to satisfy multiple tasks such as foot and CoM trajectory tracking. While our approach also leverages robot model knowledge, we avoid solving the whole-body optimization problems in real time. Instead, we train an RL policy in simulation, guided by the DCM/VRP dynamics. For comparison, we consider a state-of-the-art model-based method~\cite{egleStepTimingAdaptation2023} that determines optimal step timing and placement by solving a QP at runtime.

\begin{figure}
    \centering
\includegraphics[width=0.9\linewidth]{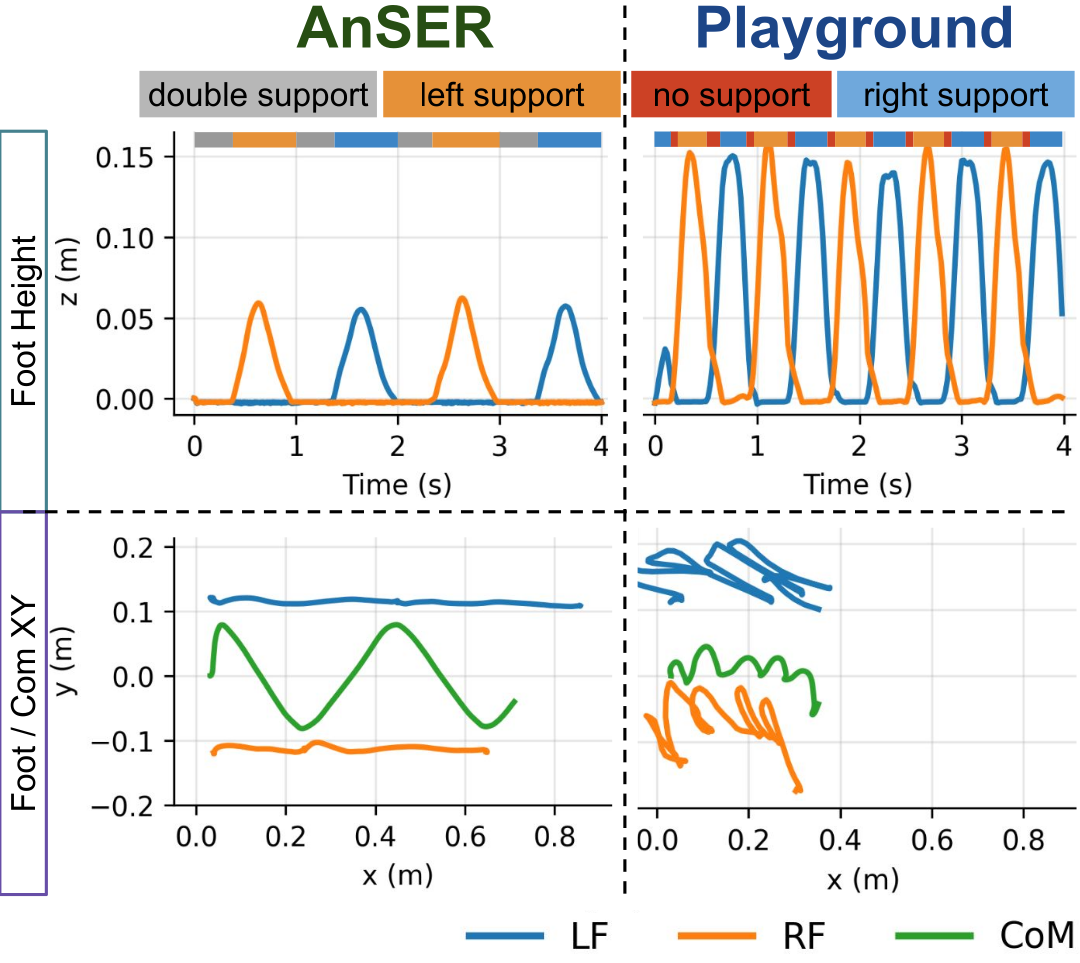}
    \caption{\textbf{Walking Analysis.} The G1 robot is commanded by a constant forward velocity of 0.2 m/s. The top-left plot shows that our \textit{AnSER} allows G1 to have a stable double stance phase, while \textit{Playground} (top right) generates fly phases when switching the stance foot. Furthermore, the robot can do slow stepping with single-stance phases of $\sim 0.5 s$ in \textit{AnSER}, requiring fine-grained GRF adjustment for balancing, while \textit{Playground} converges to a rapid stepping strategy for balancing. The bottom plots illustrate the planar trajectories of the left and right foot and the robot CoM. Our method keeps the robot strictly following the commanded velocity, while the baseline deviates significantly from it.}
    \label{fig:walking_analysis}
\end{figure}

\begin{figure*}
    \vspace{5pt}
    \centering
    \includegraphics[width=0.95\linewidth]{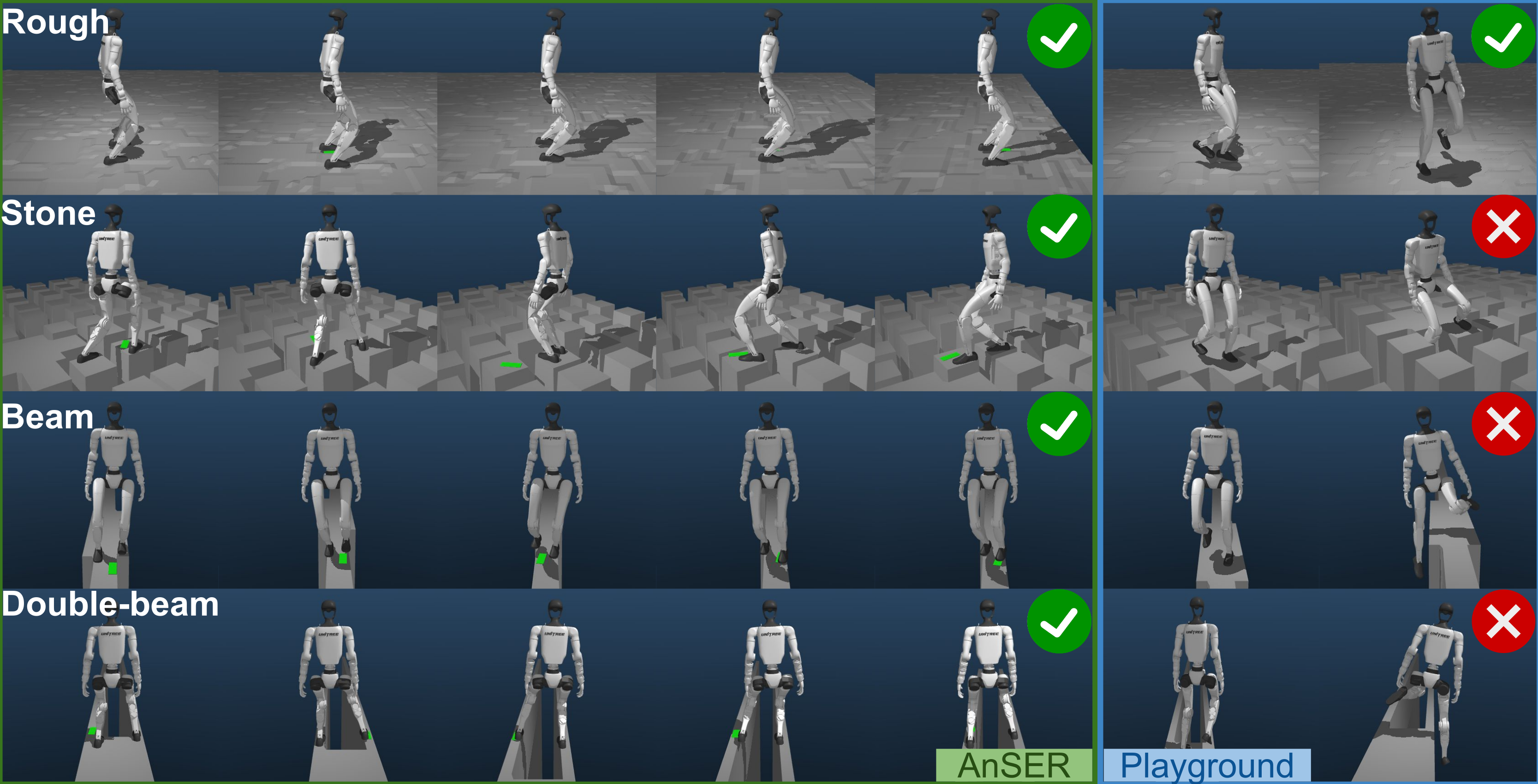}
    \caption{\textbf{Evaluation on unseen terrains.} Both \textit{AnSER} (ours) and \textit{Playground} (baseline) are trained on flat ground. Despite this, \textit{AnSER} successfully traverses unseen terrains, including randomly placed stepping stones, a narrow beam-like passage, and a double-beam passage, demonstrating superior generalization. In contrast, although the G1 robot trained with \textit{Playground} remains upright on rough terrain by taking rapid corrective steps, it exhibits large deviations from the commanded velocity and fails to traverse the narrow passages reliably.}
    \label{fig:unseen_terrains}
\end{figure*}

\begin{figure}
    \centering
    \includegraphics[width=0.99\linewidth]{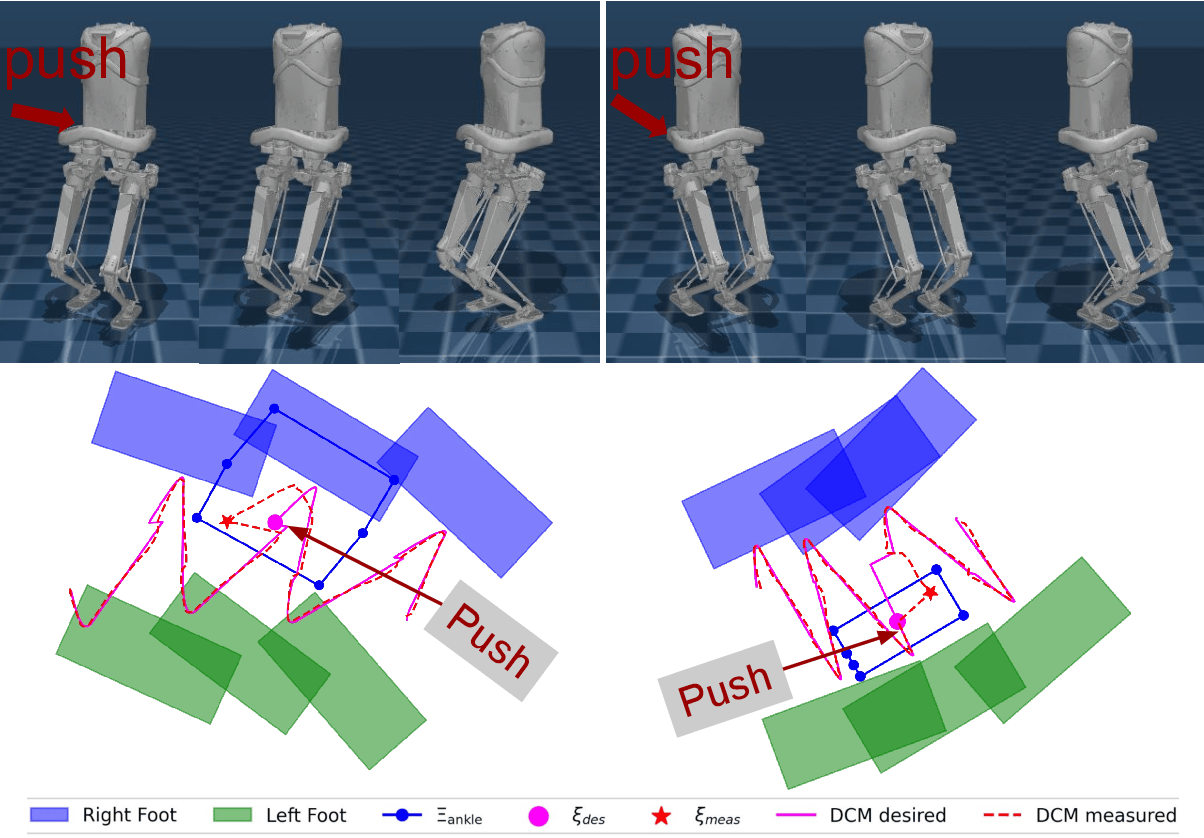}
    \caption{\textbf{Recovery without step adjustment.} Guided by the ankle strategy, our RL policy leverages foot–ground interaction to recover from disturbances without additional stepping. In the two simulations, the robot is pushed during double support (left) and single support (right). As long as $\boldsymbol{\xi}_{\mathrm{meas}}$ remains within the ankle-correctable set $\Xi_{\text{ankle}}$, the robot maintains balance solely through body control.}
    \label{fig:recover_ankle}
\end{figure}

\begin{figure*}[t]
    \vspace{5pt}
    \centering
    \includegraphics[width=0.95\linewidth]{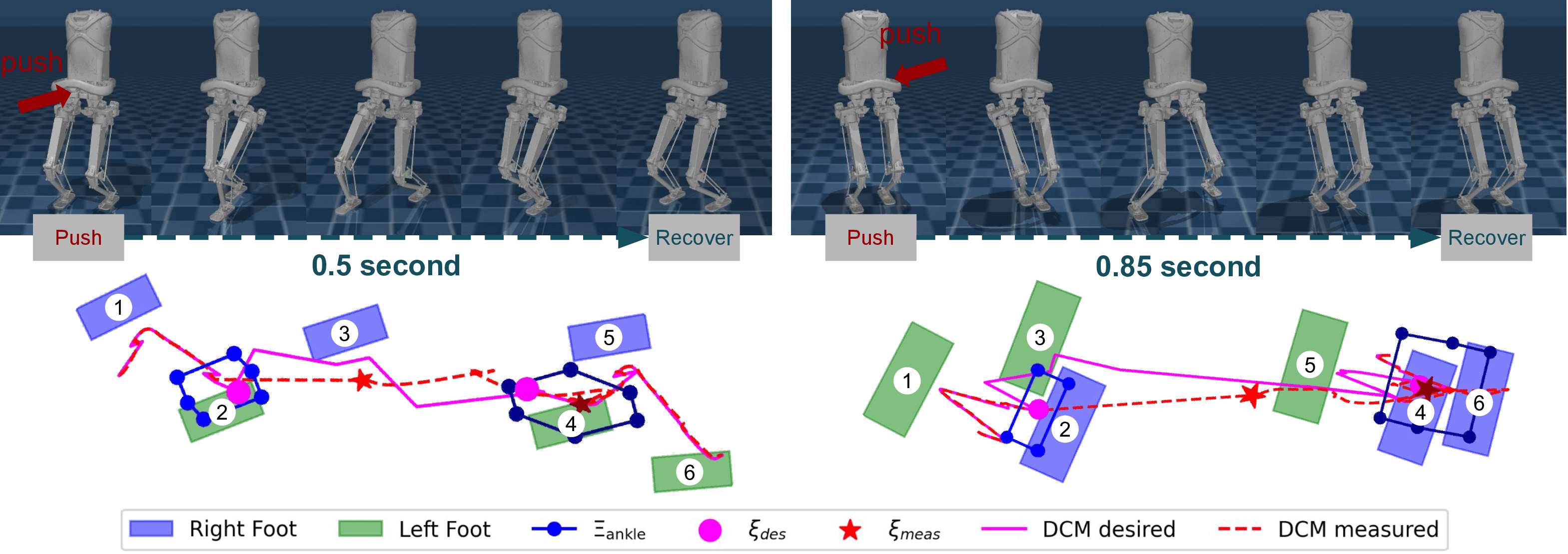}
    \caption{\textbf{Recovery with step adjustment.} When subjected to large disturbances, the robot must adapt its stepping strategy to regain balance. Guided by the stepping strategy during training, our RL policy learns to adjust step timing and location, enabling rapid recovery from strong pushes.}
    \label{fig:recover_stepping}
\end{figure*}

\subsection{Evaluation Metrics}
\label{sec:metric}
To comprehensively evaluate both bipedal walking performance and robustness against disturbances, we define the following metrics.  

\subsubsection{Metrics for Walking Performance}
\begin{itemize}
    \item \textbf{Goal Deviation (GoD)}: the Euclidean distance between the desired goal position (determined by commanded velocities) and the actual robot position (resulting from the control policy) after 10 seconds of simulated walking.  

    \item \textbf{Cost of Transport (CoT)}: the average energy cost per unit distance traveled, defined as
    \begin{equation}
        CoT = \frac{E}{mg \cdot s},
    \end{equation}
    where $E$ is the total energy consumed, $s$ is the walking distance, and $mg$ is the robot’s weight.  

    \item \textbf{DCM Tracking Error}: the average deviation between planned and measured DCM trajectories. Since the DCM encodes the center-of-mass (CoM) dynamics, accurate tracking is essential for stable and robust bipedal walking.  
\end{itemize}

\subsubsection{Metrics for Push Recovery} 

\begin{itemize}
    \item \textbf{Recovery Time}: the duration required for the robot to re-stabilize after the last push disturbance. 
    A trial is considered recovered if the velocity error falls below 0.05~m/s after the push. 

    \item \textbf{Trajectory Deviation}: the displacement of the robot base link during the period of recovery.  

    \item \textbf{Success Rate (SR)}: the percentage of trials in which the robot successfully recovers from external disturbances. For evaluation, we conduct 2000 trials with randomly sampled disturbances.
\end{itemize}

\subsection{Quantitative Evaluation}
\label{sec:quantitative}

In this section, we present a comprehensive comparison between our method and the baselines. We first evaluate bipedal walking performance, focusing on velocity tracking, energy efficiency, and gait characteristics. We then assess robustness to external disturbances.  

\subsubsection{Bipedal Walking Locomotion}
Walking performance is evaluated on two bipedal platforms, Kangaroo and G1, over 1000 simulations, each lasting 10 seconds, under varying command velocities. No external disturbances are applied in this setting.

Table~\ref{tab:quantitative} summarizes the results. The end-to-end learning baseline, Playground, deviates significantly from the desired goal position due to accumulated velocity errors over 10 seconds (i.e., 2.21 and 0.71 meters for Kangaroo and G1, respectively). In contrast, our method reduces the error to just 0.38 meters for Kangaroo and 0.17 meters for G1, essentially matching the performance of the model-based baseline. Playground also adopts a rapid stepping strategy during training to maintain balance (see Fig. \ref{fig:walking_analysis}), but this reduces energy efficiency compared to both our method and the model-based approach. Moreover, our policy achieves a comparable DCM tracking error to the model-based controller, demonstrating that a learned controller can attain the tracking accuracy of optimization-based whole-body control that exploits perfect model knowledge.

\subsubsection{Push Recovery}
To evaluate robustness, we conduct 2000 simulations, each lasting 5 seconds. An external push force is applied to the robot’s pelvis at a random time step. As in training, the disturbance consists of random forces up to 200 N in the $xy$-plane. The push is applied to the robot for a random duration from 0.1 to 0.2 seconds.

\begin{table}[]
\centering

\caption{Ablation study on $\mathcal{R}^{step}$ for G1.}
\begin{tabular}{l|ccc}
 & Recover time (s)     & Trajectory deviation (cm) & SR (\%) \\ \midrule
$\omega=0$   & 0.66  & 6.01 & 100   \\
$\omega=5$  &  0.46  & 6.77  &  100  \\
$\omega=15$  & 0.39 &  6.85 &  100  \\
$\omega=20$  & 0.34 & 6.64  &  100
\end{tabular}%

\label{tab:labtion_stepping}
\end{table}

Overall, our method, AnSER, consistently outperforms all baselines across every evaluation metric and on both robotic platforms. On average, AnSER recovers from disturbances in 0.22 s on Kangaroo and 0.34 s on G1, compared with 0.54 s and 0.49 s, respectively, for Playground. In addition, AnSER substantially reduces trajectory deviations under external pushes relative to both the model-based approach and Playground. As shown in Tab.~\ref{tab:quantitative}, AnSER reduces the trajectory deviation to less than 60\% of that achieved by Playground, while the model-based method exhibits even larger deviations from the desired trajectory.

\begin{figure*}
    \vspace{5pt}
    \centering
    \includegraphics[width=0.95\linewidth]{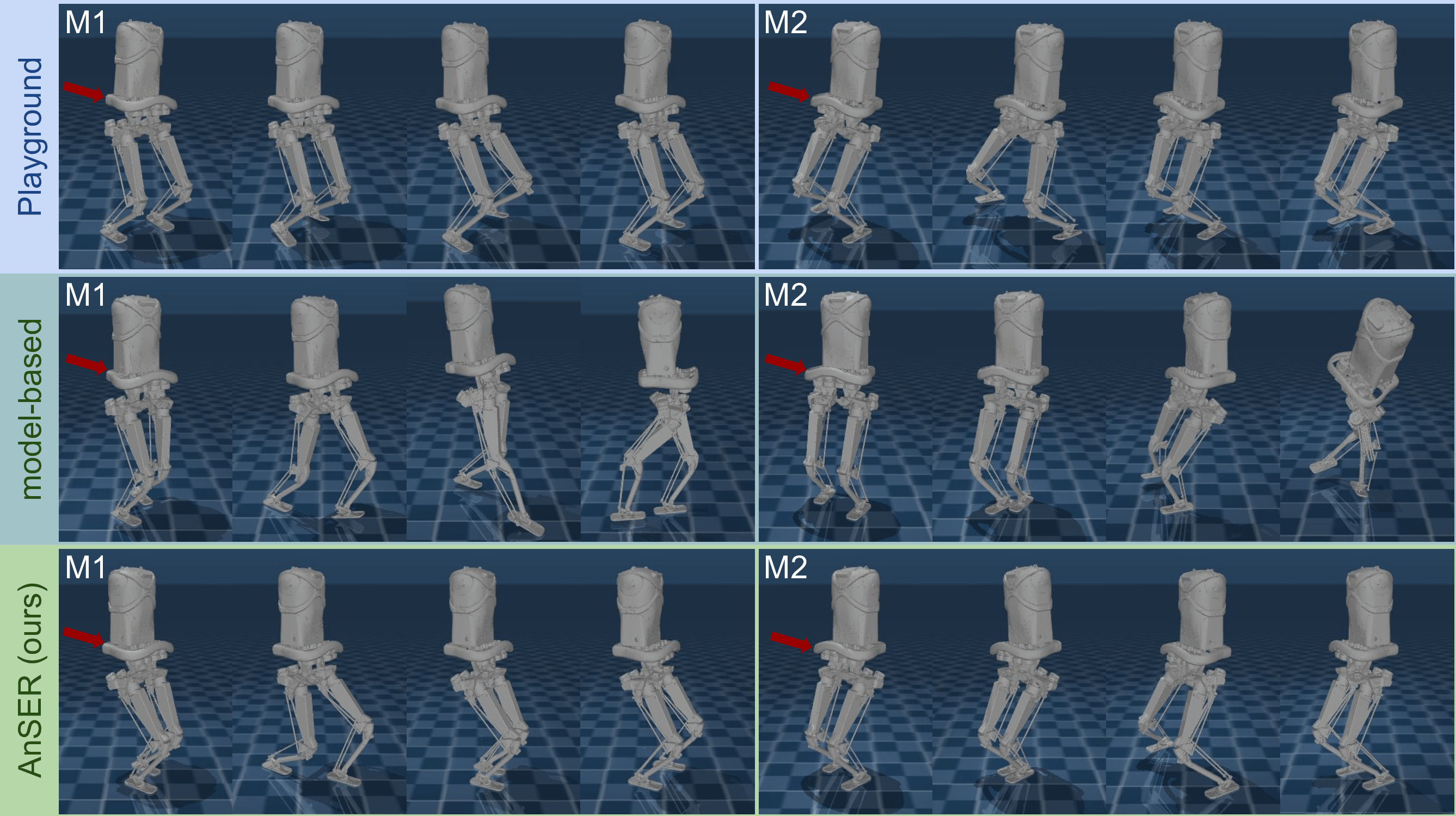}
    \caption{\textbf{Comparison with baselines.} We evaluate different methods in two simulated scenarios with large disturbances. In motion M1 (300~N push from the back), our method recovers with a single large step, whereas the Playground baseline requires multiple small steps. The model-based baseline also attempts a large step but fails because its task-space foot placement constraints do not fully capture the joint-space constraints, leading to joint-limit violations. In motion M2 (300~N push from the right), Playground briefly lifts both legs off the ground, leading to unstable contact. In contrast, our method first executes a small step to avoid self-collision, followed by a larger step that compensates for the disturbance and restores balance. Although the model-based baseline initially follows a similar strategy, its slower rate of step-time adjustment limits effectiveness.}

    \label{fig:qualitative}
\end{figure*}

Furthermore, AnSER achieves the highest recovery success rate. The model-based method performs worse than the learning-based approaches, primarily because it neglects kinematic constraints when optimizing foot placements (see Fig.~\ref{fig:qualitative}). In contrast, data-driven methods implicitly capture the robot's physical constraints, including kinematic limits, torque limits, and self-collision, from the training data, enabling more appropriate whole-body control actions during recovery.

Compared with the learning-based baseline Playground, which simply penalizes falls, AnSER promotes an optimal stepping strategy by incorporating the objective in Eq.~(\ref{eq:dcm_terminal}) as a reward-shaping term. Specifically, AnSER explicitly predicts the next footstep location and timing to drive the DCM endpoint toward convergence. This intermediate prediction also improves the interpretability of the learned policy by exposing meaningful recovery decisions that are often hidden in end-to-end black-box representations. It also provides informative feedback for subsequent planning.

\subsection{Qualitative Evaluation}
\label{sec:qualitative}

To evaluate the advantage of incorporating the ankle and stepping strategies in our method, we assess the locomotion performance under both unseen terrains and disturbances.

\subsubsection{Unseen Terrains}
During training, both AnSER and Playground are trained exclusively on flat ground with identical randomization. To evaluate their generalization capability, we test both learned policies on four previously unseen terrains: rough ground, random stepping stones, a narrow beam, and a double-beam passage. The rough terrain is generated with random height variations of up to 3cm. The stepping-stone terrain consists of discrete $30\mathrm{cm}\times30\mathrm{cm}$ platforms separated by randomly sampled gaps ranging from 9cm to 15cm. The narrow beam has a width of 10cm, whereas the double-beam terrain consists of two parallel beams separated by 32cm, each with a width of 5cm.

Since terrain perception is not included in the policy observations during training, we perform a simple terrain analysis at test time to identify feasible footholds. Specifically, we sample a $32\times32$ height map covering a $40\mathrm{cm}\times40\mathrm{cm}$ area centered on the current stance foot. For each sampled point, we evaluate the local surface to determine whether it is steppable. A location is considered steppable if all foot corners lie on a locally planar surface; otherwise, it is rejected. This heuristic generates a binary foothold mask, from which the feasible foothold closest to the desired motion command is selected.

Figure~\ref{fig:unseen_terrains} presents the performance of AnSER and Playground on the four unseen terrains. On rough terrain (top row), both methods successfully maintain locomotion under a forward velocity command. However, Playground exhibits substantial deviations from the desired trajectory, whereas AnSER closely follows the commanded motion. The remaining three terrains require precise foot placement rather than conventional high-level velocity tracking. By design, AnSER accepts both high-level velocity commands and explicit foot placement commands, enabling it to track non-periodic foothold sequences that differ significantly from the periodic velocity-tracking task used during training. Consequently, AnSER successfully traverses all challenging terrains despite never being trained on non-periodic stepping patterns. In contrast, Playground only accepts high-level velocity commands and therefore cannot exploit the available foothold information, leading to frequent failures on the discrete stepping-stone and beam terrains.

\subsubsection{Random Pushes}
Figure~\ref{fig:recover_ankle} shows two simulations where the robot is subjected to small disturbances during double support (left) and single support (right). In the left case, a 0.1~s push force of 100~N induces a 5.50~cm DCM error. Since the measured DCM $\boldsymbol{\xi}_{\mathrm{meas}}$ remains inside the convex hull defined by the ankle strategy, the robot can recover using only ground interaction. The lower plot confirms that our method successfully stabilizes without step adjustment. In the right case, during single support, the robot experiences a disturbance of 75~N for 0.1~s, causing a 4.72~cm DCM deviation. Thanks to the ankle strategy, our policy balances the robot by using the ground reaction forces.

Figure~\ref{fig:recover_stepping} illustrates robustness under large disturbances, where the ankle strategy alone is insufficient. In these scenarios, the robot must adjust its stepping to avoid falling. In the left simulation, a backward disturbance of 280~N produces a 28.71~cm DCM error. To counteract this, the robot takes two large backward steps within 0.5~s, bringing $\boldsymbol{\xi}_{\mathrm{meas}}$ back inside the convex hull, after which balance is restored through foot–ground interaction. In the right simulation, a lateral disturbance of 300~N occurs while the robot is stepping with its left leg. The robot quickly adapts by shifting the left foot laterally while avoiding self-collision, followed by a large step with the right foot that compensates for most of the disturbance momentum. After a final small adjustment, the robot successfully recovers balance.

Figure \ref{fig:qualitative} compares different methods for recovery under large disturbances in 0.2~s. In the two simulations (M1 and M2), the robot is pushed with forces of 300~N from the back and 300~N from the right, respectively. In M1, the Playground baseline executes multiple small steps to recover, whereas our method restores balance with a single large step. The model-based baseline also attempts a large step but fails because its task-space foot placement constraints (constant distance from the stance foot) do not fully capture the joint-space constraints, leading to joint-limit violations. In M2, Playground lifts both feet simultaneously, resulting in unstable ground contact. In contrast, our method first makes a quick step with the left foot to avoid self-collision, followed by a larger right-foot step that compensates for the disturbance and restores balance. Although the model-based baseline initially exhibits a comparable strategy, its slower adjustment of step timing reduces responsiveness and ultimately causes the robot to fall.

Overall, our method enables the RL policy to improve push recovery by effectively exploiting foot–ground interaction. Guided by DCM dynamics, the policy learns to generate purposeful stepping motions that rapidly counter disturbances, rather than relying on high-frequency stepping as in Playground. Unlike the model-based baseline, which optimizes step locations within a heuristic convex stepping region, our approach allows the robot to make recovery decisions while explicitly respecting its kinematic limits.

\section{Real-world Experiments}
In this section, we evaluate AnSER on the Unitree G1 humanoid robot. We first give detailed information about the hardware deployment in Sec. VI-A, followed by walking performance evaluation in Sec. VI-B. The evaluation of robustness against external disturbances is presented in Section VI-C. Finally, we demonstrate the capabilities of our controller beyond locomotion, highlighting its versatility in additional whole-body tasks in Section VI-D.

\subsection{Hardware Deployment}
To deploy our method on the Unitree G1, we implement a state estimator that fuses IMU measurements with robot kinematics to estimate the floating-base pose. In particular, we exploit the IMU measurements to estimate roll and pitch. Yaw and position are estimated using kinematics instead. For the latter, knowledge of the contact state of the robot is important. Since we have not yet implemented contact force estimation, we utilize the desired contact state for this purpose.

Known the floating-base pose, we use Pinocchio to compute the kinematic quantities required by the policy, including the CoM position, foot positions, and their orientations. During hardware experiments, the RL policy is executed at 250 Hz, while the low-level PD controller operates at 1 kHz.


\subsection{Locomotion on Hardware}
\begin{figure*}
    \centering
    \includegraphics[width=0.99\linewidth]{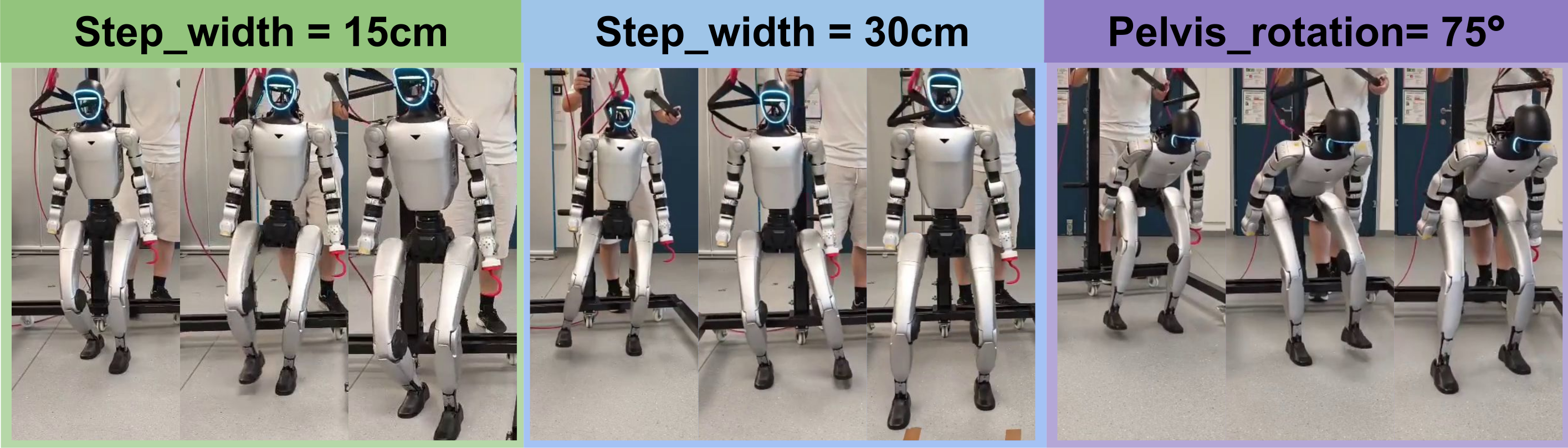}
    \caption{\textbf{Walking with diverse configurations on hardware.} Our AnSER supports diverse walking behaviors by enabling commands over gait parameters, robot height, and body orientation.}
    \label{fig:walking}
\end{figure*}

To evaluate zero-shot sim-to-real transfer of our policy, we conduct five walking experiments with a total duration of approximately 513 seconds. During these experiments, the robot was commanded with a variety of locomotion parameters, including commanded velocity, stepping frequency, foot apex height, and other gait-related commands. Table \ref{tab:tracking_g1} summarizes the tracking errors of the robot's center of mass and feet in both simulation and real-world experiments. While the CoM tracking performance degrades on the hardware compared with simulation, the foot trajectory tracking remains highly consistent between simulation and the real robot.

Figure \ref{fig:walking} illustrates the G1 robot performing walking behaviors under different configurations. Unlike conventional RL approaches, which typically command locomotion using only desired velocities, AnSER extends the command space (see Eq. \ref{eq:cmd}) with user-specified gait parameters to enable more generalized locomotion skills. Specifically, AnSER allows the robot to walk with step widths ranging from 0.1 m to 0.35 m while bending its pelvis by up to 90°. In addition, the robot's CoM height can be adjusted from 0.30 m, corresponding to a low-squatting posture, to 0.65 m, corresponding to a fully upright stance. These expressive locomotion configurations not only improve the robot's traversability over challenging terrain (see Fig. \ref{fig:unseen_terrains}) but also significantly enlarge the robot's reachable workspace, facilitating more capable high-level manipulation tasks.


\begin{table}[]
\centering
\caption{\textbf{Comparison of tracking errors between simulation and hardware.} }

\begin{tabular}{l|ccc}
                  & CoM (cm)     & Left Foot (cm)    & Right Foot (cm)   \\ \midrule
Simulation             &  0.45   &   0.65   & 0.66 \\      
Hardware             & 1.46     & 0.81      & 0.70         
\end{tabular}%

\label{tab:tracking_g1}
\end{table}

\subsection{Robustness on Hardware}
\begin{figure*}
    \centering
    \includegraphics[width=0.99\linewidth]{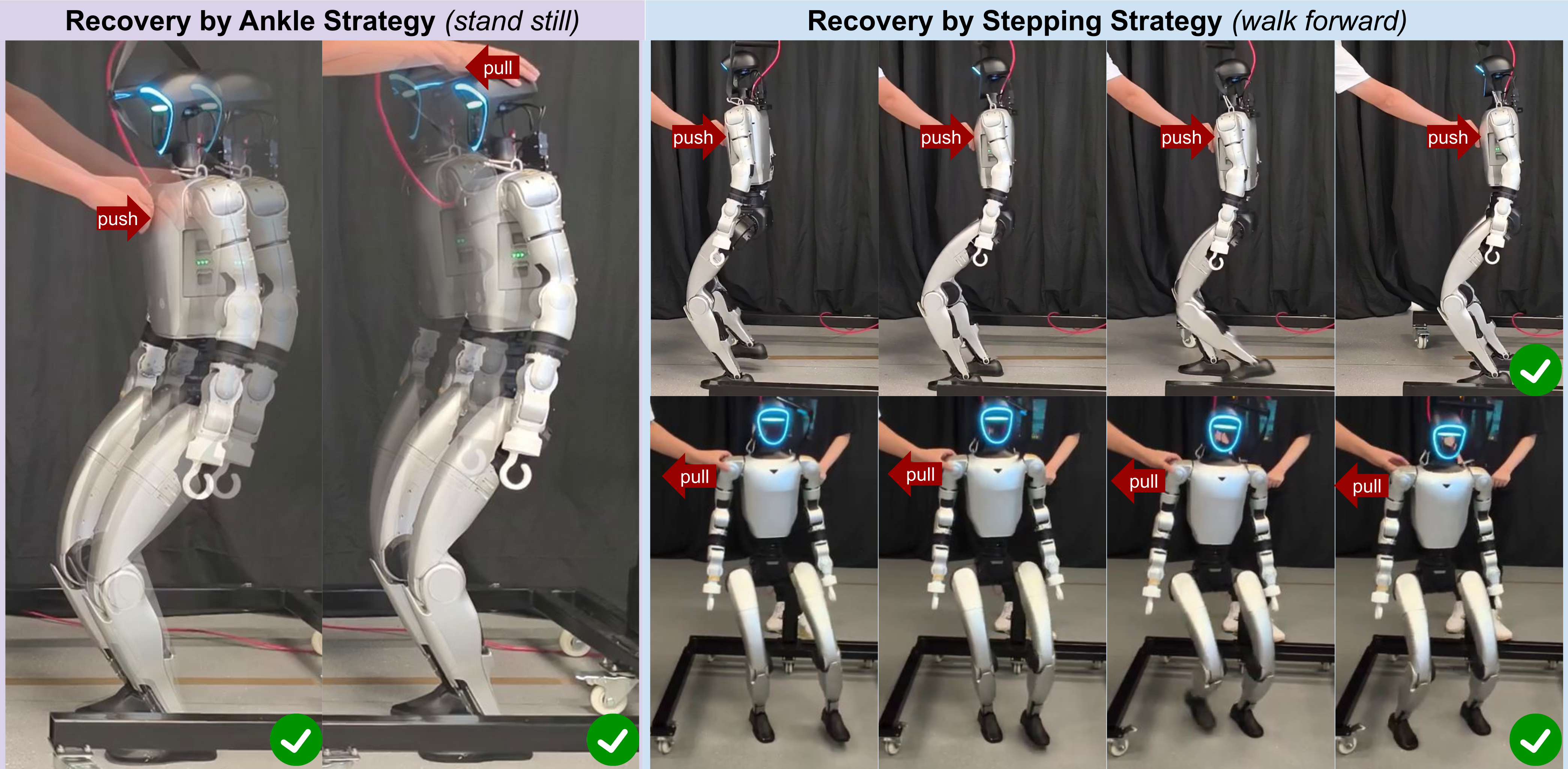}
    \caption{\textbf{Push recovery on hardware.} Under moderate disturbances (left), the G1 robot employs an ankle strategy to resist external pushes by exploiting ground reaction forces, avoiding additional steps. For larger disturbances, our policy adapts the robot's stepping strategy to maintain balance (right). In the top row, when pushed forward, the G1 temporarily steps backward to recover its balance. In the bottom row, the robot is pulled to the right during right-foot single support. Instead of following the nominal stepping plan, the policy places the left foot to the right and rapidly extends the right leg farther outward, enabling the robot to recover from the lateral disturbance.}
    \label{fig:push_recovery}
\end{figure*}
\begin{figure*}
    \centering
    \includegraphics[width=0.99\linewidth]{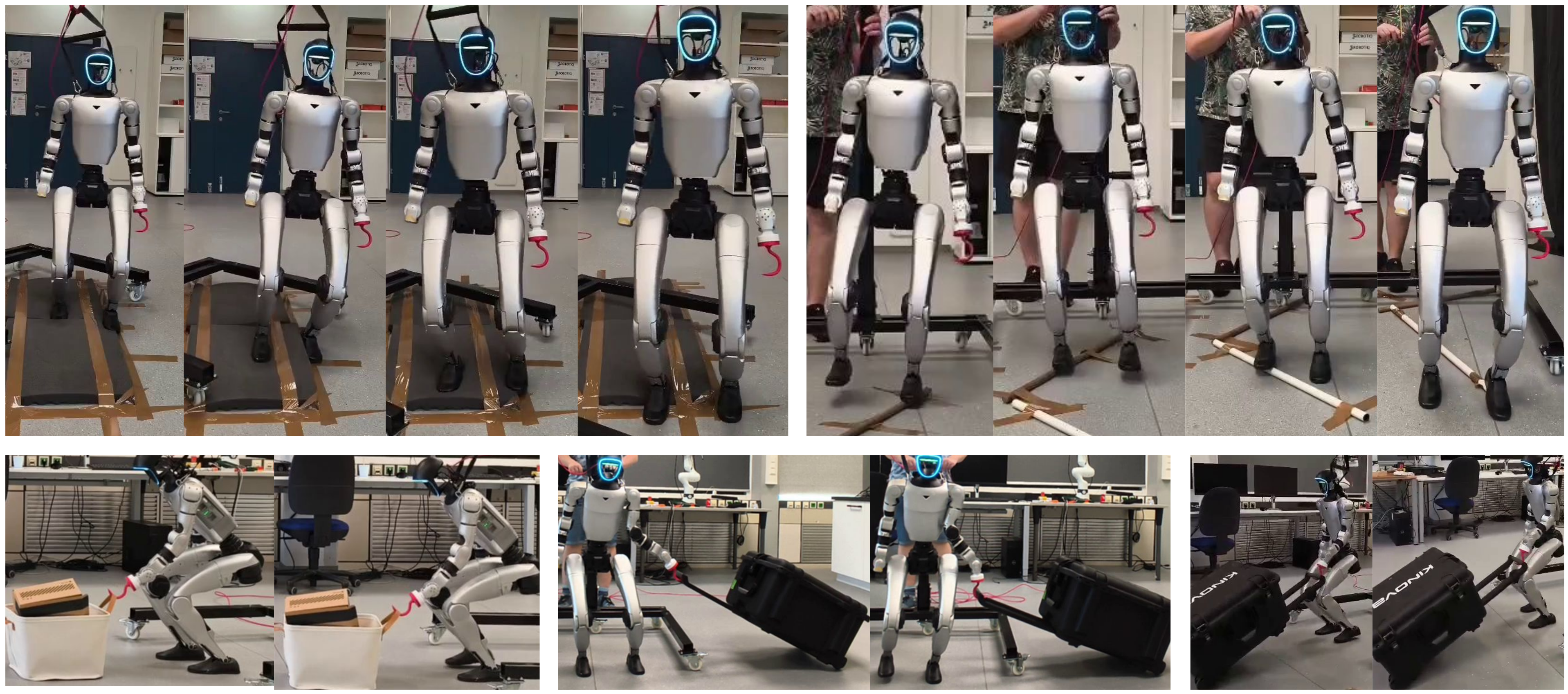}
    \caption{\textbf{Hardware evaluation on unseen disturbances.} The policy is trained on a flat surface, where the robot is subjected to external push disturbances and does not interact with any objects. We evaluate the robot's locomotion on challenging scenarios, including stepping on a soft mat (top left) and walking over round sticks (top right). In the bottom figure, the robot is required to pull objects while maintaining stable locomotion.}
    \label{fig:unseen_disturbances}
\end{figure*}
AnSER enhances the robot's balancing capability by encouraging the exploitation of ground reaction forces (GRFs) and optimizing stepping parameters. To achieve this, AnSER focuses on regulating the robot's CoM motion based on analytical DCM/VRP dynamics. Through the application of ARC, the policy learns to actively generate appropriate GRFs for balance control, which is crucial for stable locomotion on deformable terrains such as soft mats. Furthermore, leveraging GRF-based balance control enables the robot to perform stable low-frequency stepping, allowing it to generate sustained interaction forces for pulling heavy objects instead of relying on rapid stepping adjustments.

Figure \ref{fig:push_recovery} illustrates the application of the ankle and stepping strategies on the G1 robot for recovery from external pushes. Using the ankle strategy, the robot remains stationary and counteracts the pushing forces through GRFs, as shown on the left. The experiments aslo demonstrate that AnSER adapts the robot stepping strategy to recover from large pushes while walking forward. Under backward pushes, the G1 robot temporarily steps backward to regain balance (top right). Under a lateral push to the right while the left foot is in swing, the robot places its left foot to the right and rapidly extends its right leg outward to recover its balance.

Figure \ref{fig:unseen_disturbances} presents experiments on the G1 robot under disturbances that are unseen during training. In the top-left experiment, the G1 robot walks on a soft mat with a slow stepping frequency, demonstrating the effectiveness of GRF exploitation in AnSER. In the top-right experiment, the robot traverses rollable sticks with diameters of approximately 2 cm. As the robot walks, the rolling motion of the sticks introduces additional disturbances to the robot's balance. In the bottom experiment, the G1 robot drags objects while walking. Despite pulling a heavy suitcase during both forward and lateral locomotion, the robot does not require additional stepping adjustments; instead, it maintains stability by exploiting GRFs for balancing.

\subsection{General Whole-body Control}
\begin{figure*}
    \centering
    \includegraphics[width=0.99\linewidth]{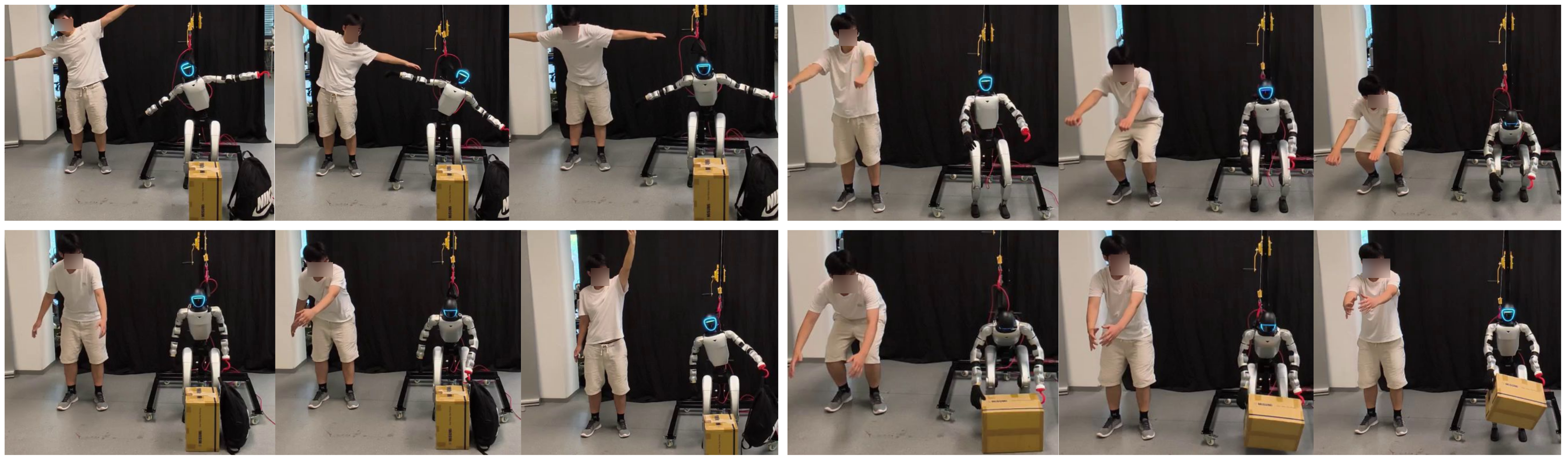}
    \caption{\textbf{Teleoperation for unseen upper-body motions.} During training, no motion commands are applied to the upper-body joints. Nevertheless, the learned policy can accommodate arbitrary upper-body movements by maintaining balance through center-of-mass feedback. Furthermore, we demonstrate the integration of AnSER with teleoperation to perform simple manipulation tasks, such as picking up objects from the floor, while preserving stable whole-body balance.}
    \label{fig:teleop}
\end{figure*}

\begin{figure}
    \centering
    \includegraphics[width=0.99\linewidth]{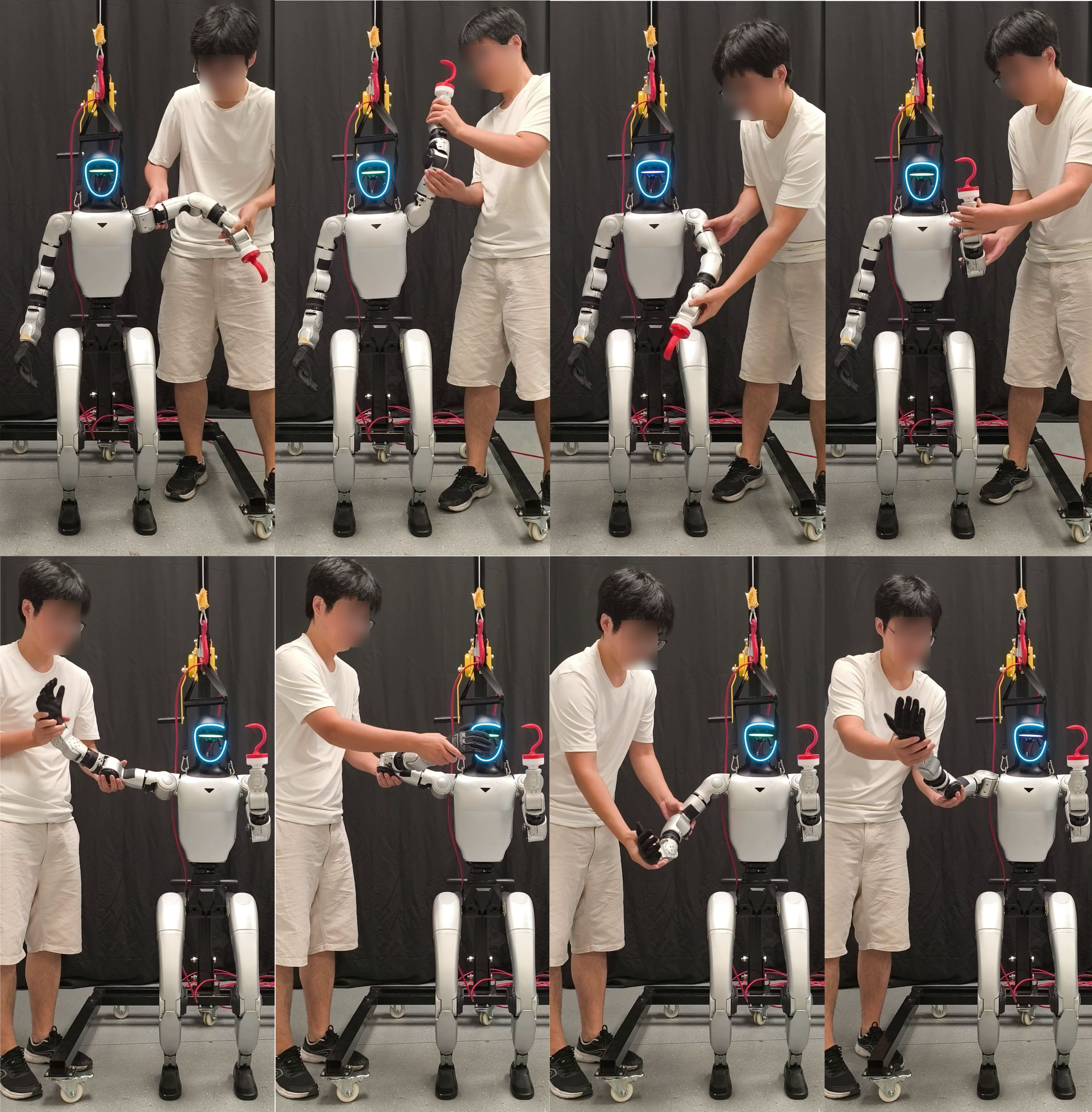}
    \caption{\textbf{Kinesthetic teaching upper-body motions.} By \textit{AnSER}, users are able to demonstrate arbitrary robot upper-body motions by direct physical interaction, which can be seamlessly connected with learning-from-demonstration methods.}
    \label{fig:kinethetic}
\end{figure}
Conventional RL-based methods for locomotion and loco-manipulation typically observe the robot's full-body joint states, such as all 29 joints of the G1 robot. However, relying on full-body observations limits their generalization to unseen motions, as the policy must experience and explore the corresponding commanded motions during training in order to reproduce them during inference. In contrast, AnSER addresses this limitation by using only the leg joint states (12 joints for G1) and the robot's CoM, while intentionally excluding upper-body joint observations. Since upper-body joints are not part of the policy observation space, unseen upper-body motions do not appear as out-of-distribution inputs to the learned policy. Nevertheless, upper-body motions influence the robot's CoM, which remains observable by the policy. By training the policy to regulate CoM dynamics for balance control, AnSER can inherently compensate for disturbances induced by unseen upper-body movements.

Figure \ref{fig:teleop} presents experiments integrating AnSER with a vision-based motion retargeting framework. Specifically, an Azure Kinect camera is used to estimate human 3D poses, and ImitationNet \cite{ImitationNet} is employed to retarget human motions into robot joint configurations. During teleoperation, the human pelvis height and orientation are mapped to the robot's pelvis height and orientation, while the human upper-body pose is transferred to the G1's 17 upper-body joints. AnSER successfully maintains balance despite the disturbances generated by upper-body motions.

Additionally, we evaluate AnSER in manipulation scenarios, where the robot is teleoperated to pick up a backpack and a carton from the floor. These experiments demonstrate that AnSER can serve as a robust low-level locomotion controller for high-level manipulation tasks without requiring task-specific training or extensive exposure to manipulation behaviors during learning.

Furthermore, \textit{AnSER} naturally supports diverse upper-body control interfaces while autonomously maintaining lower-body balance. As illustrated in Fig.~\ref{fig:kinethetic}, this capability establishes a direct connection between \textit{AnSER} and humanoid kinesthetic teaching \cite{6697020}, where users physically interact with the robot to demonstrate desired behaviors. As one of the most widely adopted paradigms in learning from demonstration for robotic manipulators, kinesthetic teaching has proven highly effective for efficiently acquiring manipulation skills. By excluding upper-body joints from the inputs to the lower-body balance controller, \textit{AnSER} allows users to directly teach or command motions across the full upper-body joint space while the controller continuously maintains whole-body stability by adjusting the CoM movements. This design has the potential to significantly accelerate skill acquisition for humanoid robots, combining the efficiency of kinesthetic teaching with the mobility and terrain adaptability afforded by bipedal locomotion.

\section{Discussion and Future Work}
Robust bipedal locomotion under external disturbances has been extensively studied using both model-based and learning-based approaches. Model-based methods leverage knowledge of robot dynamics to optimize task-space objectives and compute joint torques through whole-body control. However, these approaches often simplify or neglect hard constraints, such as self-collision avoidance, kinematic limits, and other physical constraints. Furthermore, real-world uncertainties, including unknown friction coefficients, damping properties, and link mass variations, are difficult to accurately model, resulting in degraded performance when deployed on real hardware.

In contrast, learning-based approaches typically reduce reliance on explicit models and train policies through trial-and-error interactions. However, end-to-end learning methods are often sensitive to reward design, which can lead to suboptimal behaviors or limited generalization. Meanwhile, incorporating prior model knowledge, such as foot support regions and CoM dynamics, remains crucial for effectively regulating GRFs and improving locomotion stability and performance.

Our method, AnSER, combines the strengths of both paradigms by integrating learning-based flexibility with model-based insights. By leveraging reduced yet essential model knowledge through analytical DCM/VRP dynamics, AnSER provides principled guidance for policy learning. As a result, the robot learns to exploit ground interactions for robust balance control without unnecessary stepping, while simultaneously developing effective stepping strategies for rapid recovery from large disturbances. Through its carefully designed observation space, AnSER achieves strong generalization to unseen disturbances arising from challenging terrains and upper-body motions. Furthermore, AnSER demonstrates the potential to serve as a robust low-level locomotion controller for high-level manipulation tasks by decoupling locomotion stability from manipulation learning, thereby reducing the complexity of developing manipulation capabilities for legged robots.

In future work, we aim to extend AnSER toward high-impact interaction tasks, where robots must continuously generate large forces to push, pull, or manipulate heavy objects. Such tasks require precise modulation of ground reaction forces and coordinated adjustment of body posture. Therefore, we plan to incorporate VRP-based control and explicit body pose regulation into AnSER to further enhance its capability for dynamic whole-body interactions.

\section{Conclusion}
We introduced \emph{AnSER}, a reinforcement learning framework for robust bipedal locomotion that unifies task-space planning and whole-body control within a single policy. The key novelty of AnSER lies in incorporating reduced yet essential model knowledge to guide end-to-end policy learning, enabling the robot to effectively exploit its physical capabilities for robust balance and recovery. To this end, we proposed to use the Ankle Recovery Condition (ARC), derived from DCM and VRP dynamics, which provides a principled criterion for evaluating recoverability through ankle motion and center-of-pressure (CoP) modulation. During training, ARC encourages the policy to exploit ground reaction forces (GRFs) for balance control, thereby reducing unnecessary stepping. When stepping becomes necessary for recovery, the policy learns to optimize foot placement and timing, enabling rapid disturbance rejection and convergence to the desired trajectory.

Extensive simulation and hardware experiments demonstrate that AnSER outperforms both model-based and purely end-to-end learning baselines in robust locomotion under external disturbances. Furthermore, through its carefully designed observation space, AnSER naturally decouples locomotion stabilization from upper-body manipulation. Since upper-body joint states are not included in the policy input, arbitrary upper-body motions can be commanded during deployment without requiring explicit training demonstrations. These motions influence the robot's center of mass (CoM), which remains observable and regulated by AnSER, allowing the robot to maintain balance while performing whole-body manipulation tasks.

These findings highlight the potential of combining reduced-order analytical models with reinforcement learning to achieve adaptive, robust, and versatile humanoid locomotion, providing a scalable foundation for integrating locomotion with higher-level interaction and manipulation capabilities.

\bibliographystyle{IEEEtran}
\bibliography{bib/relatedworks, bib/ICRA2026}

@inproceedings{bangRLaugmentedMPCFramework2024,
  title = {{{RL-augmented MPC Framework}} for {{Agile}} and {{Robust Bipedal Footstep Locomotion Planning}} and {{Control}}},
  booktitle = {2024 {{IEEE-RAS}} 23rd {{International Conference}} on {{Humanoid Robots}} ({{Humanoids}})},
  author = {Bang, Seung Hyeon and Jov{\'e}, Carlos Arribalzaga and Sentis, Luis},
  year = {2024},
  month = nov,
  pages = {607--614},
  issn = {2164-0580},
  doi = {10.1109/Humanoids58906.2024.10769914},
  urldate = {2025-09-13}
}

@inproceedings{chenReinforcementLearningReducedorder2024,
  title = {Reinforcement {{Learning}} for {{Reduced-order Models}} of {{Legged Robots}}},
  booktitle = {2024 {{IEEE International Conference}} on {{Robotics}} and {{Automation}} ({{ICRA}})},
  author = {Chen, Yu-Ming and Bui, Hien and Posa, Michael},
  year = {2024},
  month = may,
  pages = {5801--5807},
  doi = {10.1109/ICRA57147.2024.10610747},
  urldate = {2025-09-13}
}

@inproceedings{kamiokaSimultaneousOptimizationZMP2018,
  title = {Simultaneous {{Optimization}} of {{ZMP}} and {{Footsteps Based}} on the {{Analytical Solution}} of {{Divergent Component}} of {{Motion}}},
  booktitle = {Proc. {{IEEE Int}}. {{Conf}}. {{Robot}}. {{Automat}}.},
  author = {Kamioka, Takumi and Kaneko, Hiroyuki and Takenaka, Toru and Yoshiike, Takahide},
  year = 2018,
  pages = {1763--1770},
  issn = {2577-087X},
  doi = {10.1109/ICRA.2018.8460572}
}

@inproceedings{diedamOnlineWalkingGait2008,
  title = {Online Walking Gait Generation with Adaptive Foot Positioning through {{Linear Model Predictive}} Control},
  booktitle = {Proc. {{IEEE}}/{{RSJ Int}}. {{Conf}}. {{Intell}}. {{Robots Syst}}.},
  author = {Diedam, Holger and Dimitrov, Dimitar and Wieber, Pierre-Brice and Mombaur, Katja and Diehl, Moritz},
  year = 2008,
  pages = {1121--1126},
  issn = {2153-0866},
  doi = {10.1109/IROS.2008.4651055}
}

@inproceedings{romualdiOnlineNonlinearCentroidal2022a,
  title = {Online {{Non-linear Centroidal MPC}} for {{Humanoid Robot Locomotion}} with {{Step Adjustment}}},
  booktitle = {2022 {{International Conference}} on {{Robotics}} and {{Automation}} ({{ICRA}})},
  author = {Romualdi, Giulio and Dafarra, Stefano and L'Erario, Giuseppe and Sorrentino, Ines and Traversaro, Silvio and Pucci, Daniele},
  year = 2022,
  month = may,
  pages = {10412--10419},
  doi = {10.1109/ICRA46639.2022.9811670},
  urldate = {2026-07-28}
}

@inproceedings{duanSimtoRealLearningFootstepConstrained2022a,
  title = {Sim-to-{{Real Learning}} of {{Footstep-Constrained Bipedal Dynamic Walking}}},
  booktitle = {2022 {{International Conference}} on {{Robotics}} and {{Automation}} ({{ICRA}})},
  author = {Duan, Helei and Malik, Ashish and Dao, Jeremy and Saxena, Aseem and Green, Kevin and Siekmann, Jonah and Fern, Alan and Hurst, Jonathan},
  year = {2022},
  month = may,
  pages = {10428--10434},
  doi = {10.1109/ICRA46639.2022.9812015},
  urldate = {2025-09-13}
}

@misc{egleEnhancingModelBasedStep2024,
  title = {Enhancing {{Model-Based Step Adaptation}} for {{Push Recovery}} through {{Reinforcement Learning}} of {{Step Timing}} and {{Region}}},
  author = {Egle, Tobias and Yan, Yashuai and Lee, Dongheui and Ott, Christian},
  year = {2024},
  month = nov,
  number = {arXiv:2411.01000},
  eprint = {2411.01000},
  publisher = {arXiv},
  doi = {10.48550/arXiv.2411.01000},
  urldate = {2024-11-07},
  archiveprefix = {arXiv}
}

@inproceedings{egleStepTimingAdaptation2023,
  title = {Step and {{Timing Adaptation}} during {{Online DCM Trajectory Generation}} for {{Robust Humanoid Walking}} with {{Double Support Phases}}},
  booktitle = {{{IEEE-RAS}} 22nd {{Int}}. {{Conf}}. {{Humanoid Robots}}},
  author = {Egle, Tobias and Englsberger, Johannes and Ott, Christian},
  year = {2023},
  issn = {2164-0580},
  doi = {10.1109/Humanoids57100.2023.10375152},
  urldate = {2024-07-12}
}

@article{englsbergerThreeDimensionalBipedalWalking2015,
  title = {Three-{{Dimensional Bipedal Walking Control Based}} on {{Divergent Component}} of {{Motion}}},
  author = {Englsberger, Johannes and Ott, Christian and {Albu-Sch{\"a}ffer}, Alin},
  year = {2015},
  journal = {IEEE Trans. Robot.},
  issn = {1941-0468},
  doi = {10.1109/TRO.2015.2405592}
}

@inproceedings{englsbergerTorqueBasedDynamicWalking2018,
  title = {Torque-{{Based Dynamic Walking}} - {{A Long Way}} from {{Simulation}} to {{Experiment}}},
  booktitle = {2018 {{IEEE International Conference}} on {{Robotics}} and {{Automation}} ({{ICRA}})},
  author = {Englsberger, Johannes and Mesesan, George and Werner, Alexander and Ott, Christian},
  year = {2018},
  month = may,
  pages = {440--447},
  issn = {2577-087X},
  doi = {10.1109/ICRA.2018.8462862}
}

@inproceedings{kajita3DLinearInverted2001,
  title = {The {{3D}} Linear Inverted Pendulum Mode: A Simple Modeling for a Biped Walking Pattern Generation},
  shorttitle = {The {{3D}} Linear Inverted Pendulum Mode},
  booktitle = {Proc. {{IEEE Int}}. {{Conf}}. {{Robot}}. {{Automat}}.},
  author = {Kajita, S. and Kanehiro, F. and Kaneko, K. and Yokoi, K. and Hirukawa, H.},
  year = {2001},
  volume = {1},
  pages = {239--246},
  doi = {10.1109/IROS.2001.973365}
}

@article{mesesanConvexPropertiesCenterofMass2018,
  title = {Convex {{Properties}} of {{Center-of-Mass Trajectories}} for {{Locomotion Based}} on {{Divergent Component}} of {{Motion}}},
  author = {Mesesan, George and Englsberger, Johannes and Ott, Christian and {Albu-Sch{\"a}ffer}, Alin},
  year = {2018},
  journal = {IEEE Robotics and Automation Letters},
  volume = {3},
  number = {4},
  pages = {3449--3456},
  issn = {2377-3766},
  doi = {10.1109/LRA.2018.2853557}
}

@inproceedings{mesesanDynamicWalkingCompliant2019,
  title = {Dynamic {{Walking}} on {{Compliant}} and {{Uneven Terrain}} Using {{DCM}} and {{Passivity-based Whole-body Control}}},
  booktitle = {Proc. {{IEEE-RAS}} 19th {{Int}}. {{Conf}}. {{Humanoid Robots}}},
  author = {Mesesan, George and Englsberger, Johannes and Garofalo, Gianluca and Ott, Christian and {Albu-Sch{\"a}ffer}, Alin},
  year = {2019},
  pages = {25--32},
  issn = {2164-0580},
  doi = {10.1109/Humanoids43949.2019.9035053}
}

@inproceedings{mesesanOnlineDCMTrajectory2021,
  title = {Online {{DCM Trajectory Adaptation}} for {{Push}} and {{Stumble Recovery}} during {{Humanoid Locomotion}}},
  booktitle = {Proc. {{IEEE Int}}. {{Conf}}. {{Robot}}. {{Automat}}.},
  author = {Mesesan, George and Englsberger, Johannes and Ott, Christian},
  year = {2021},
  issn = {2577-087X},
  doi = {10.1109/ICRA48506.2021.9560808}
}

@inproceedings{prattCapturePointStep2006,
  title = {Capture {{Point}}: {{A Step}} toward {{Humanoid Push Recovery}}},
  shorttitle = {Capture {{Point}}},
  booktitle = {Proc. {{IEEE-RAS}} 15th {{Int}}. {{Conf}}. {{Humanoid Robots}}},
  author = {Pratt, Jerry and Carff, John and Drakunov, Sergey and Goswami, Ambarish},
  year = {2006},
  pages = {200--207},
  issn = {2164-0580},
  doi = {10.1109/ICHR.2006.321385}
}

@inproceedings{roigHardwareDesignControl2022,
  title = {On the {{Hardware Design}} and {{Control Architecture}} of the {{Humanoid Robot Kangaroo}}},
  booktitle = {6th {{Workshop}} on {{Legged Robots}} during the {{Int}}. {{Conf}}. {{Robot}}. {{Automat}}.},
  author = {Roig, Adria and Kothakota, Sai Kishor and Miguel, Narcis and Fernbach, Pierre and Hoffman, Enrico Mingo and Marchionni, Luca},
  year = {2022},
  urldate = {2023-07-10},
  langid = {english}
}

@inproceedings{sugiharaRealtimeHumanoidMotion2002,
  title = {Real-Time Humanoid Motion Generation through {{ZMP}} Manipulation Based on Inverted Pendulum Control},
  booktitle = {Proc. {{IEEE Int}}. {{Conf}}. {{Robot}}. {{Automat}}.},
  author = {Sugihara, T. and Nakamura, Y. and Inoue, H.},
  year = {2002},
  volume = {2},
  pages = {1404--1409},
  doi = {10.1109/ROBOT.2002.1014740}
}

@inproceedings{takenakaRealTimeMotion2009,
  title = {Real Time Motion Generation and Control for Biped Robot -1st Report: {{Walking}} Gait Pattern Generation-},
  shorttitle = {Real Time Motion Generation and Control for Biped Robot -1st Report},
  booktitle = {Proc. {{IEEE}}/{{RSJ Int}}. {{Conf}}. {{Intell}}. {{Robots Syst}}.},
  author = {Takenaka, Toru and Matsumoto, Takashi and Yoshiike, Takahide},
  year = {2009},
  issn = {2153-0866},
  doi = {10.1109/IROS.2009.5354662}
}

@article{wangLearning3DBipedal2023,
  title = {Learning {{3D Bipedal Walking}} with {{Planned Footsteps}} and {{Fourier Series Periodic Gait Planning}}},
  author = {Wang, Song and Piao, Songhao and Leng, Xiaokun and He, Zhicheng},
  year = {2023},
  month = jan,
  journal = {Sensors},
  volume = {23},
  number = {4},
  pages = {1873},
  publisher = {Multidisciplinary Digital Publishing Institute},
  issn = {1424-8220},
  doi = {10.3390/s23041873},
  urldate = {2025-09-13},
  copyright = {http://creativecommons.org/licenses/by/3.0/},
  langid = {english}
}

@inproceedings{wieberTrajectoryFreeLinear2006,
  title = {Trajectory {{Free Linear Model Predictive Control}} for {{Stable Walking}} in the {{Presence}} of {{Strong Perturbations}}},
  booktitle = {2006 6th {{IEEE-RAS International Conference}} on {{Humanoid Robots}}},
  author = {Wieber, Pierre-brice},
  year = {2006},
  month = dec,
  pages = {137--142},
  issn = {2164-0580},
  doi = {10.1109/ICHR.2006.321375}
}

@misc{zakkaMuJoCoPlayground2025,
  title = {{{MuJoCo Playground}}},
  author = {Zakka, Kevin and Tabanpour, Baruch and Liao, Qiayuan and Haiderbhai, Mustafa and Holt, Samuel and Luo, Jing Yuan and Allshire, Arthur and Frey, Erik and Sreenath, Koushil and Kahrs, Lueder A. and Sferrazza, Carmelo and Tassa, Yuval and Abbeel, Pieter},
  year = {2025},
  month = feb,
  number = {arXiv:2502.08844},
  eprint = {2502.08844},
  primaryclass = {cs},
  publisher = {arXiv},
  doi = {10.48550/arXiv.2502.08844},
  urldate = {2025-09-14},
  archiveprefix = {arXiv},
  langid = {english}
}

@article{
ANYmalparkour,
author = {David Hoeller  and Nikita Rudin  and Dhionis Sako  and Marco Hutter },
title = {ANYmal parkour: Learning agile navigation for quadrupedal robots},
journal = {Science Robotics},
year = {2024},
}

@inproceedings{
rudin2021learning,
title={Learning to Walk in Minutes Using Massively Parallel Deep Reinforcement Learning},
author={Nikita Rudin and David Hoeller and Philipp Reist and Marco Hutter},
booktitle={5th Annual Conference on Robot Learning },
year={2021},
url={https://openreview.net/forum?id=wK2fDDJ5VcF}
}

@ARTICLE{ictrl,
  author={Yan, Yashuai and Mascaro, Esteve Valls and Egle, Tobias and Lee, Dongheui},
  journal={IEEE Robotics \& Automation Magazine}, 
  title={I-CTRL: Imitation to Control Humanoid Robots Through Bounded Residual Reinforcement Learning}, 
  year={2025},
  volume={32},
  number={1},
  pages={59-67},
  doi={10.1109/MRA.2025.3527284}}

@INPROCEEDINGS{ImitationNet,
  author={Yan, Yashuai and Mascaro, Esteve Valls and Lee, Dongheui},
  booktitle={2023 IEEE-RAS 22nd International Conference on Humanoid Robots (Humanoids)}, 
  title={ImitationNet: Unsupervised Human-to-Robot Motion Retargeting via Shared Latent Space}, 
  year={2023},}

@inproceedings{todorov2012mujoco,
  title={MuJoCo: A physics engine for model-based control},
  author={Todorov, Emanuel and Erez, Tom and Tassa, Yuval},
  booktitle={2012 IEEE/RSJ International Conference on Intelligent Robots and Systems},
  pages={5026--5033},
  year={2012},
  organization={IEEE},
  doi={10.1109/IROS.2012.6386109}
}

@misc{fernandez2024stable,
      title={Fast and Accurate Deep Network Learning by Exponential Linear Units (ELUs)}, 
      author={Djork-Arné Clevert and Thomas Unterthiner and Sepp Hochreiter},
      year={2016},
      eprint={1511.07289},
      archivePrefix={arXiv},
      primaryClass={cs.LG},
      url={https://arxiv.org/abs/1511.07289}, 
}

@INPROCEEDINGS{9561705,
  author={Duan, Helei and Dao, Jeremy and Green, Kevin and Apgar, Taylor and Fern, Alan and Hurst, Jonathan},
  booktitle={2021 IEEE International Conference on Robotics and Automation (ICRA)}, 
  title={Learning Task Space Actions for Bipedal Locomotion}, 
  year={2021},
  volume={},
  number={},}

@ARTICLE{11395590,
  author={Yan, Yashuai and Egle, Tobias and Ott, Christian and Lee, Dongheui},
  journal={IEEE Robotics and Automation Letters}, 
  title={Efficiently Learning Robust Torque-Based Locomotion Through Reinforcement With Model-Based Supervision}, 
  year={2026}}

@ARTICLE{11346966,
  author={Li, Kejun and Olkin, Zachary and Yue, Yisong and Ames, Aaron D.},
  journal={IEEE Robotics and Automation Letters}, 
  title={CLF-RL: Control Lyapunov Function Guided Reinforcement Learning}, 
  year={2026},}

@misc{araujo2025retargetingmattersgeneralmotion,
      title={Retargeting Matters: General Motion Retargeting for Humanoid Motion Tracking}, 
      author={Joao Pedro Araujo and Yanjie Ze and Pei Xu and Jiajun Wu and C. Karen Liu},
      year={2025},
      eprint={2510.02252},
      archivePrefix={arXiv},
      primaryClass={cs.RO},
      url={https://arxiv.org/abs/2510.02252}, 
}

@article{yang2025omniretarget,
  title={Omniretarget: Interaction-preserving data generation for humanoid whole-body loco-manipulation and scene interaction},
  author={Yang, Lujie and Huang, Xiaoyu and Wu, Zhen and Kanazawa, Angjoo and Abbeel, Pieter and Sferrazza, Carmelo and Liu, C Karen and Duan, Rocky and Shi, Guanya},
  booktitle={2026 IEEE International Conference on Robotics and Automation (ICRA)}, 
}

@ARTICLE{11397669,
  author={Yan, Yashuai and Lee, Dongheui},
  journal={IEEE Robotics \& Automation Magazine}, 
  title={Cross-Embodiment Imitation: Learning a Unified Latent Space for Multirobot Control}, 
  year={2026},}

@misc{ze2025twist2scalableportableholistic,
      title={TWIST2: Scalable, Portable, and Holistic Humanoid Data Collection System}, 
      author={Yanjie Ze and Siheng Zhao and Weizhuo Wang and Angjoo Kanazawa and Rocky Duan and Pieter Abbeel and Guanya Shi and Jiajun Wu and C. Karen Liu},
      booktitle={2026 IEEE International Conference on Robotics and Automation (ICRA)}
}

@InProceedings{
	ze2025twist,
	title = {TWIST: Teleoperated Whole-Body Imitation System},
	author = {Ze, Yanjie and Chen, Zixuan and Araujo, Joao Pedro and Cao, Zi-ang and Peng, Xue Bin and Wu, Jiajun and Liu, Karen},
	booktitle = {Proceedings of The 9th Conference on Robot Learning},
	year = {2025},
	publisher = {PMLR},
}

@misc{liao2025beyondmimicmotiontrackingversatile,
      title={BeyondMimic: From Motion Tracking to Versatile Humanoid Control via Guided Diffusion}, 
      author={Qiayuan Liao and Takara E. Truong and Xiaoyu Huang and Yuman Gao and Guy Tevet and Koushil Sreenath and C. Karen Liu},
      year={2025},
      eprint={2508.08241},
      archivePrefix={arXiv},
      primaryClass={cs.RO},
      url={https://arxiv.org/abs/2508.08241}, 
}

@misc{margolis2025softmimiclearningcompliantwholebody,
      title={SoftMimic: Learning Compliant Whole-body Control from Examples}, 
      author={Gabriel B. Margolis and Michelle Wang and Nolan Fey and Pulkit Agrawal},
      year={2025},
      eprint={2510.17792},
      archivePrefix={arXiv},
      primaryClass={cs.RO},
      url={https://arxiv.org/abs/2510.17792}, 
}

@INPROCEEDINGS{6697020,
  author={Ott, Christian and Henze, Bernd and Lee, Dongheui},
  booktitle={2013 IEEE/RSJ International Conference on Intelligent Robots and Systems}, 
  title={Kinesthetic teaching of humanoid motion based on whole-body compliance control with interaction-aware balancing}}

\vfill

\end{document}